\documentclass[lettersize,journal]{IEEEtran}
\usepackage{amsmath,amsfonts}

\usepackage{algorithm}
\usepackage{array}
\usepackage[caption=false,font=normalsize,labelfont=sf,textfont=sf]{subfig}
\usepackage{textcomp}
\usepackage{stfloats}
\usepackage{url}
\usepackage{verbatim}
\usepackage{graphicx}
\usepackage{booktabs}
\usepackage{multirow}
\usepackage{cite}
\usepackage{algpseudocode}
\usepackage{pifont}
\usepackage{xcolor}

\usepackage{makecell}

\begin{document}

\title{CRIP: Channel Level Representation Injection for Personalized One-Shot Federated Learning}

% \author{IEEE Publication Technology,~\IEEEmembership{Staff,~IEEE,}
%         % <-this % stops a space
% \thanks{This paper was produced by the IEEE Publication Technology Group. They are in Piscataway, NJ.}% <-this % stops a space
% \thanks{Manuscript received April 19, 2021; revised August 16, 2021.}}
\author{
Zijian Jiang,
Chaoli Sun,
Handing Wang,
and Xilu Wang%
\thanks{Zijian Jiang is with the School of
Computer Science and Electronic Engineering, University of Surrey,
Guildford, Surrey GU2 7XH, U.K.}%
\thanks{Chaoli Sun is with the School of Computer Science and Technology,
Taiyuan University of Science and Technology, Taiyuan 030024, China.}%
\thanks{Handing Wang is with the School of Artificial Intelligence,
Xidian University, Xi'an 710071, China.}%
\thanks{Xilu Wang is with the Department of Computer Science,
University of Surrey, Guildford, U.K.
(e-mail: xiluwang@surrey.ac.uk).}%
\thanks{Corresponding author: Xilu Wang.}%
}
% The paper headers
\markboth{Journal of \LaTeX\ Class Files,~Vol.~14, No.~8, August~2021}%
{Shell \MakeLowercase{\textit{et al.}}: A Sample Article Using IEEEtran.cls for IEEE Journals}

%\IEEEpubid{0000--0000/00\$00.00~\copyright~2021 IEEE}
% Remember, if you use this you must call \IEEEpubidadjcol in the second
% column for its text to clear the IEEEpubid mark.

\maketitle

\begin{abstract}
One-shot federated learning (OSFL) has emerged as a promising collaborative model learning framework with only a single round of communication, offering significant advantages in communication efficiency and privacy preservation. However, OSFL often faces inherent limitations under severe domain heterogeneity across clients due to the lack of iterative knowledge exchange. Most existing OSFL methods require an auxiliary public dataset for knowledge distillation or leverage statistical information for parameter-level aggregation, overlooking feature shift caused by domain heterogeneity. To address these challenges, we propose CRIP, a personalized OSFL framework that operates in the representation space via channel-level feature alignment. To achieve this, each client uploads its feature extractor to the server, which broadcasts all extractors back to every client. Since not all source clients share compatible feature distributions with the target client, indiscriminate fusion of cross-client features would introduce domain-specific noise. Therefore, CRIP effectively measures the channel-wise representational similarity between the target client and each source client on a small local mini-batch, and selectively fuses only the most compatible features. Extensive experiments on domain-heterogeneous benchmarks such as DomainNet, PACS, and Office-Home demonstrate that CRIP consistently outperforms local models and state-of-the-art baselines, validating the effectiveness of representation-space personalization under extreme domain heterogeneity. %CRIP works in representation space instead of parameter space. Feature fusion in this space helps preserve semantic information. It also helps suppress domain-specific residual noise. These properties cannot be guaranteed by parameter-space aggregation. To keep the method efficient, CRIP uses a small set of domain-specific samples. Each client applies all client models to its local data and extracts channel-wise feature representations. It then integrates useful cross-client information for prediction on its own domain. The method requires only a single communication round, in which feature extractors are uploaded to and downloaded from the server. Extensive experiments on domain-heterogeneous benchmarks such as DomainNet, PACS, and Office-Home demonstrate that CRIP consistently outperforms local models and state-of-the-art baselines, validating the effectiveness of representation-space personalization under extreme domain heterogeneity.
\end{abstract}

\begin{IEEEkeywords}
One-shot Federated Learning, Representation learning, Personalized Federated Learning, Domain Heterogeneity.
\end{IEEEkeywords}
\section{Introduction}
\begin{figure}[tbp!]
  \begin{center}
    \centerline{\includegraphics[width=\columnwidth]{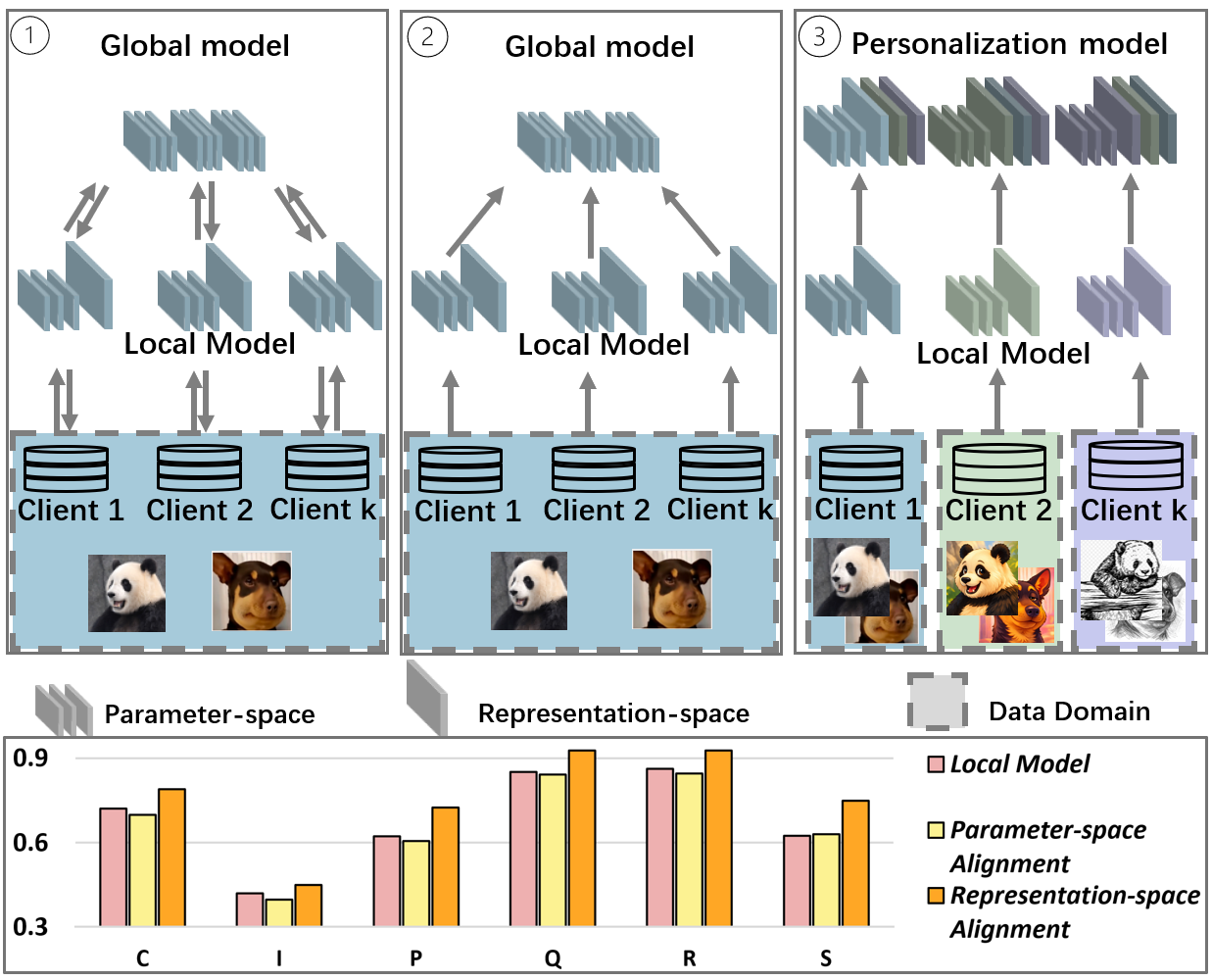}}
    \caption{FL settings and representation vs. parameter space alignment. The upper row illustrates three FL scenarios: (i) Multi-round FL with homogeneous domains across clients, (ii) One-shot FL with homogeneous domains, and (iii) One-shot FL with heterogeneous domains (ours). The lower panel compares local model performance, parameter-space alignment, and representation-space alignment under domain-heterogeneous OSFL.}
    \label{R-P}
  \end{center}
\end{figure}

\IEEEPARstart{F}{ederated} learning (FL) is a widely used decentralized machine learning (ML) paradigm. It allows multiple clients to collaboratively train models without exposing their local private data~\cite{FLsurvey}. In general, this is achieved through iterative parameter exchange across multiple communication rounds. A central server coordinates this process, as shown in Fig. \ref{R-P} \textcircled{1}. However, iterative communication introduces substantial overhead. It also increases the risk of privacy leakage and attacks. In addition, it requires strict transmission synchronization \cite{talpinifedbens,qi2025fedtmos,allouah2024revisiting}. One-shot federated learning (OSFL) has emerged as an alternative paradigm. It enables collaborative learning with only a single communication round, as shown in Fig. \ref{R-P} \textcircled{2}. This design substantially reduces communication overhead and privacy-related costs\cite{OSFL}. In OSFL, each client independently trains a local model and sends it to the server. The server then aggregates these local models into a global model. No iterative updates or feedback are required \cite{zhang2022dense,heinbaugh2023data,guan2025capture}.

Existing OSFL methods can be broadly categorized into three groups:
(i) knowledge distillation methods using a public auxiliary dataset, such as FedD3~\cite{FedD3}, Co-Boosting~\cite{Co-Boosting}, and FOL~\cite{FOL};
(ii) generative methods by leveraging generative models (e.g., diffusion models) to synthesize data for knowledge transfer, including FedBiP~\cite{DBLP:journals/corr/abs-2410-04810}, FedDEO~\cite{Feddeo}, and FGL~\cite{FGL}; and
(iii) approaches that extract and aggregate statistical quantities from local models directly in the parameter space, such as FedLPA~\cite{liu2024fedlpa}, FedFisher~\cite{jhunjhunwala2024fedfisher}, and FuseFL~\cite{tang2024fusefl}. Although these methods reduce communication cost to a single round, they face a fundamental challenge. With limited communication, models cannot adequately align or exchange information across clients, which undermines effective knowledge transfer. Furthermore, many existing approaches rely on impractical assumptions, such as access to public or high-quality synthetic data tailored to each client’s domain. Meanwhile, methods based on parameter-space alignment becomes unreliable under strong client heterogeneity, leading to degraded accuracy compared with iterative FL~\cite{zengdoes}.

These challenges become more pronounced when client data are domain-heterogeneous. Here, domain heterogeneity refers to the variation in data style and distribution across clients, which is often caused by differences in acquisition devices, collection protocols, or preprocessing pipelines. %In this setting, there are no iterative communication rounds to gradually realign different local models \cite{zengdoes}.
As illustrated in Fig.~\ref{R-P} \textcircled{3}, medical institutions may use different scanners and acquisition protocols, inducing significant domain shifts~\cite{MedicalImaging}. In such settings, independently trained models can drift far apart in parameter space. This makes parameter-space alignment inherently unreliable~\cite{Fed2}. Moreover, with only a single aggregation step, OSFL may fail to correct mismatches in cross-client transfer, as no later rounds are available for refinement.

We conduct an empirical study on DomainNet~\cite{DomainNet}, comparing parameter-based and feature-based model aggregation (see the lower panel of Fig.~\ref{R-P}). We observe that, interestingly, representation-space alignment yields substantially more effective cross-domain knowledge transfer than parameter-space methods~\cite{Fed2}. We hypothesize that parameters become entangled with domain-specific information after independent local training. In contrast, representations retain more transferable semantic structure. As a result, representation alignment is less sensitive to domain shifts. See Section \ref{sec:theory} for further analysis. These observations motivate us to tackle domain-heterogeneous OSFL from a representation-space perspective. 

Several studies have explored representation-space information in FL, including Fed$^2$~\cite{Fed2}, pFedAFM~\cite{pFedAFM}, and FedRDA~\cite{FedRDA}. However, these methods generally require multi-round iterative updates to gradually align or stabilize representations. This makes them incompatible with the one-shot setting. To address this gap, we propose CRIP, a channel-level representation injection framework for personalized OSFL. 
CRIP uses feature extractors from other client models to process the target client’s local data. It then aligns the source features with the target features. After that, it fuses the aligned source features with the target client’s own features through residual injection. Without iterative communication or parameter updates, this process supports personalized adaptation for each local domain by improving performance on the target client domain.

Our contributions are summarized as follows:
\begin{itemize}
    \item %We propose CRIP, a personalized OSFL framework for domain-heterogeneous settings. Under single-round communication, domain heterogeneity severely limits effective cross-client knowledge transfer. We further show, both empirically and theoretically, that CRIP outperforms parameter-space methods under domain heterogeneity.
    We identify domain heterogeneity as a critical yet underexplored challenge in OSFL, and reveal through empirical analysis that feature-space aggregation yields superior cross-client generalization compared to parameter-space counterparts under domain heterogeneity. Motivated by this, we propose CRIP, a personalized OSFL framework that exploits channel-level representations to capture transferable feature information across clients with heterogeneous domains.
    \item To achieve representation-level aggregation, we introduce a channel-level representation injection mechanism for cross-client knowledge transfer. It works in representation space instead of parameter space by injecting semantically aligned source features as residuals while keeping local parameters unchanged.
    \item We demonstrate state-of-the-art performance on multiple domain-heterogeneous benchmarks, consistently outperforming competing OSFL and personalized FL baselines.  We further show both empirically and theoretically that CRIP outperforms parameter-space methods under domain heterogeneity.
\end{itemize}

\section{Related Works}
\subsection{One-Shot Federated Learning}
Existing OSFL methods generally integrate information by updating model weights (e.g., via distillation objectives) or aggregating parameter-space statistics. Prior methods can be grouped into three classes based on how cross-client information is introduced under this constraint.

Knowledge distillation methods introduce a public or auxiliary dataset and perform parameter-level or output-level distillation of multiple local models on this data. Representative approaches include FedD3~\cite{FedD3}, FedSD2C~\cite{zhang2024one}, DOSFL~\cite{zhou2020arXiv}, Co-Boosting~\cite{Co-Boosting}, and FOL~\cite{FOL}. While effective, these methods explicitly depend on public data, and their performance is highly sensitive to the quality and representativeness of such data.

Generative methods leverage powerful generative models, such as diffusion models, to synthesize data that replaces real public datasets for knowledge distillation, including FedBiP~\cite{DBLP:journals/corr/abs-2410-04810}, XorMixFL~\cite{shin2020xor}, FedPFT~\cite{beitollahi2024parametric}, OSGAN~\cite{kasturi2023osgan}, FedDEO~\cite{Feddeo}, FedSD2C~\cite{zhang2024one}, and FGL~\cite{FGL}. Although this strategy alleviates the need for real auxiliary data, knowledge transfer is still realized through parameter updates or distillation objectives and often incurs substantial computational overhead.

Statistical methods do not rely on explicit data. Instead, they infer statistical quantities or structural information from local model parameters and aggregate them once on the server side. Typical examples include FedLPA~\cite{liu2024fedlpa}, FedFisher~\cite{jhunjhunwala2024fedfisher}, and FuseFL~\cite{tang2024fusefl}. These methods are fully data-free. However, they work only in the parameter space. As a result, they generally struggle to explicitly model feature distribution shifts across domain-heterogeneous clients.

Despite promising results, existing OSFL methods face significant challenges under domain heterogeneity. When clients train on data from different domains, their models diverge sharply in parameter space. This makes parameter-based alignment or aggregation fundamentally limited. To our knowledge, no prior work integrates knowledge directly in representation space. This space may preserve more meaningful semantic correspondences across domains, despite divergence in parameter space.

\subsection{Representation Based Federated Learning}
Representation-based FL has been widely studied as a solution to mitigate heterogeneity by aligning feature spaces across clients. Most of them are proposed for the standard multi-round FL setting, where repeated communication enables representations to be progressively calibrated through iterative local optimization and global aggregation \cite{shi2024clip,zhang2023gpfl,chen2023fraug}. For example, Fed$^2$~\cite{Fed2} enforces feature alignment via grouped convolutions and gradient redirection, but requires architectural uniformity across clients and iterative updates to stabilize alignment.
FedFA \cite{zhou2023fedfa} and FedFM~\cite{fedfm} introduce class anchors in the feature space and guide local learning through multi-round iterations. Subspace projection methods quantify representation discrepancies through geometric transformations. FedFed~\cite{yang2023fedfed} distills feature information to encourage consistency across clients, relying on repeated rounds for distilled signals to accumulate and stabilize. FedRDA~\cite{FedRDA} projects local features onto a global principal feature subspace, using the orthogonal residual as a measure of deviation. While offering geometric interpretability, this approach depends on iterative updates to jointly minimize distribution shift. 

Feature-mixing methods dynamically fuse global and local knowledge. GPFL~\cite{zhang2023gpfl} simultaneously learns global and personalized feature information to balance generalization and personalization. pFedAFM~\cite{pFedAFM} adaptively mixes representations from a global feature extractor and heterogeneous local models using learnable weight vectors. However, such mixing is typically applied at a coarse granularity (e.g., batch- or sample-level), which can miss finer channel-level semantic mismatches between domains. Recent work improves representation robustness under more challenging regimes by introducing stronger priors or augmentation strategies. CLIP-guided federated learning leverages vision--language priors to cope with heterogeneous and long-tailed distributions \cite{shi2024clip}. FRAug addresses non-IID features through representation augmentation \cite{chen2023fraug}.

Despite their effectiveness, these methods are designed for multi-round training and require learning additional components or parameter updates to achieve representation alignment. They are fundamentally incompatible with one-shot settings, where iterative calibration is unavailable.

\section{Methodology}
\begin{figure*}[ht]
  \vskip 0.2in
  \begin{center}
    \centerline{\includegraphics[width=\textwidth]{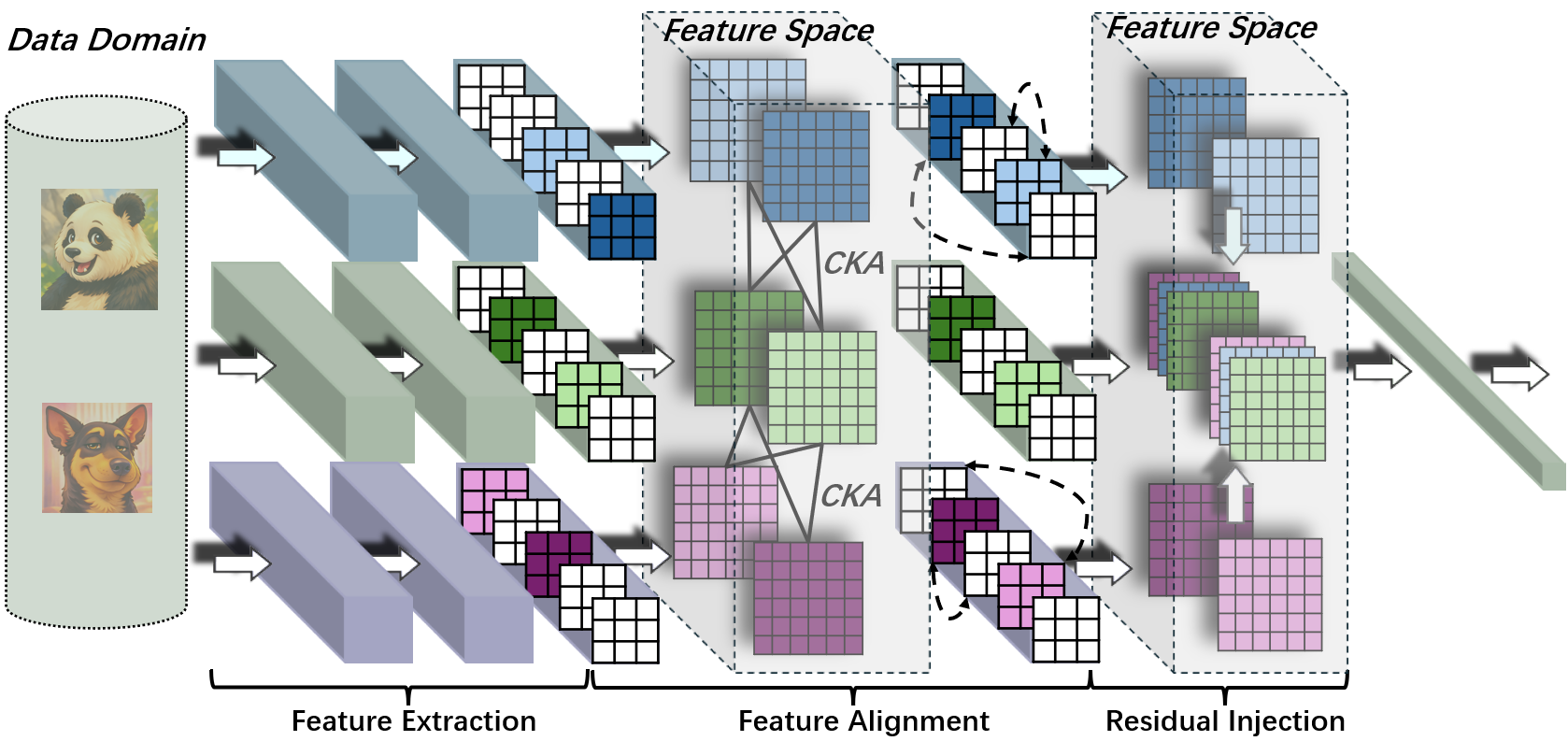}}
    \caption{Overview of CRIP. The target local model (in green) and client models (in other colors) perform feature extraction to obtain feature maps. In the representation space, we compute channel-level similarity using CKA to establish channel correspondences across models. Aligned channel outputs from source models are integrated into the local model's activations via residual injection, enabling knowledge transfer without parameter modification.}
    \label{Fig:Overview of CRIP}
  \end{center}
\end{figure*}

\subsection{Problem formulation}
\label{sec:problem}

We consider OSFL with $K$ clients under domain heterogeneity. Each client $k \in \{1, \ldots, K\}$ communicates only once and holds local data $\mathcal{D}_k$ from a distinct domain with a distribution $\mathbb{P}_k(x,y)$, where $x$ denotes the input and $y$ the label. Domain heterogeneity is denoted by $\mathbb{P}_k(x,y) \neq \mathbb{P}_{k'}(x,y)$ for $k \neq k'$. Each client $k$ trains a model $f(\cdot;\theta_k)$ on $\mathcal{D}_k$, decomposed into a feature extractor and a prediction head:
\begin{equation}
f(x;\theta_k)=h_k\!\left(g_k(x;\theta_k)\right),
\end{equation}
where $g_k(\cdot;\theta_k)$ denotes the feature extractor and $h_k(\cdot)$ the prediction head. Our goal is to perform personalized OSFL to improve each client's local model performance by leveraging knowledge from other clients under the one-shot constraint. Instead of aggregating or updating parameters, we formulate personalization as a representation-space transformation applied to intermediate features.%Let $f(\cdot;\theta_k)$ denote the model trained by client $k$ on $\mathcal{D}_k$.

For an input $x$, each local model $k$ produces an intermediate representation at layer $\ell$,
\begin{equation}
\label{FE}
\mathcal{Z}^{(\ell)}_k(x) = g_k(x;\theta_k).
\end{equation}
Given representations $\{\mathcal{Z}_k(x)\}_{k=1}^K$ from all clients, we seek a client-specific operator $\mathcal{T}_{k}^{(\ell)}$ that constructs a personalized feature map
\begin{equation}
\hat{z}_k^{(\ell)}(x)=\mathcal{T}_{k}^{(\ell)}\!\left(\{z_i^{(\ell)}(x)\}_{i=1}^{K}\right),
\end{equation}
subject to: (i) $\mathcal{T}_{k}^{(\ell)}$ operates only on representations, (ii) introduces no learnable parameters, and (iii) performs no updates to $\{\theta_k\}_{k=1}^K$. The personalized prediction for client $k$ is obtained by applying its original head:
\begin{equation}
\hat{y}_k = h_k\!\left(\hat{z}_k^{(\ell)}(x)\right),
\end{equation}
which remains fixed throughout. In the following section, we construct $\mathcal{T}_{k}^{(\ell)}$ via training-free channel-level feature alignment.

\subsection{Residual Injection via Feature Alignment}
The proposed CRIP is a representation-space OSFL framework that transfers cross-client knowledge through channel-level feature alignment and residual injection. It does not aggregate model parameters. Motivated by privacy concerns, in CRIP, each client uploads only the feature extractor of its locally trained model to the server in a single communication round, rather than the entire model. The server then broadcasts the collected feature extractors back to all clients. As illustrated in Fig.~\ref{Fig:Overview of CRIP}, each client then personalizes its model through three steps: feature extraction, channel alignment, and residual injection. We describe this personalization process for a target client $k$. Each client independently applies this procedure to construct its own personalized model.
\paragraph{Feature extraction}
Given an input $x$, let $z_k^{(\ell)}(x)$ denote the feature map at layer $\ell$ that consists of feature maps from different channels:
\begin{equation}
z_k^{(\ell)}(x)
= \{\, z_{k,c}^{(\ell)}(x) \,\}_{c=1}^{C_\ell},
\end{equation}
where $z_{k,c}^{(\ell)}(x)$ denotes the feature map of the $c$-th channel. To extract channel level representations, we perform a mini batch forward pass using $\mathcal{B}=\{x_1,\dots,x_B\}$ sampled from the local data of the target client $k$.
All remaining client feature extractors forward the same mini batch $\mathcal{B}$ during calibration.

For each channel $c$ and sample $x_b\in\mathcal{B}$, we flatten the spatial response:
\begin{equation}
\phi_{k,c}^{(\ell)}(x_b)
=\mathrm{vec}\!\left(z_{k,c}^{(\ell)}(x_b)\right),
\end{equation}
where $\mathrm{vec}(\cdot)$ denotes vectorization over the spatial dimensions.
We then concatenate the flattened responses across samples to obtain a channel specific representation:
\begin{equation}
\psi_{k,c}^{(\ell)}(\mathcal{B})
=\mathrm{Concat}\big(\phi_{k,c}^{(\ell)}(x_1),\dots,\phi_{k,c}^{(\ell)}(x_B)\big).
\end{equation}
Collecting all channels yields the set of channel wise representations
\begin{equation}
\mathcal{Z}_k^{(\ell)}(\mathcal{B})
=
\{\psi_{k,c}^{(\ell)}(\mathcal{B})\}_{c=1}^{C_\ell},
\end{equation}
which serves as channel level representations in the subsequent alignment stage.

\paragraph{Feature alignment}
Given the channel wise representations $\mathcal{Z}_k^{(\ell)}(\mathcal{B})$, CRIP identifies transferable channels via channel-level alignment across clients. For a target client $k$ and another source client $i \in \mathcal{N}_k$ with $ \mathcal{N}_k= \{1,\ldots,K\}\setminus\{k\}$ denoting all source clients, we measure the channel-wise representation similarity between client $k$ and client $i$ at layer $\ell$ using centered kernel alignment (CKA) \cite{kornblith2019similarity}.

Specifically, for channel $c$ of client $k$ and channel $c'$ of client $i$, the alignment score is defined as
\begin{equation}
\label{CKAselect}
s_{k,i}^{(\ell)}(c,c')
=
\mathrm{CKA}\!\left(
\psi_{k,c}^{(\ell)}(\mathcal{B}),
\psi_{i,c'}^{(\ell)}(\mathcal{B})
\right).
\end{equation}

For each target channel $c$, we select the most similar channel from each source client $i$ by maximizing the alignment score. For each source client $i \in \mathcal{N}_k$, let $c_i^{*}$ denote its aligned channel for the target channel $c$ (as determined in the alignment step). This establishes a channel correspondence between each target channel and one channel from each source client, forming the aligned channel set $\mathcal{A}_{k,c}^{(\ell)} = \{(i, c_i^{*}) \mid i \in \mathcal{N}_k\}$  that captures channels with consistent activation patterns across heterogeneous domains.

\paragraph{Residual injection}
The aligned channel set $\mathcal{A}_{k,c}^{(\ell)}$ is determined once using the calibration batch $\mathcal{B}$ and remains fixed thereafter. Based on this established correspondence, residual injection is performed independently for each input during inference. 

For an arbitrary input $x$, let $z_{k,c}^{(\ell)}(x)$ denote the original feature map of channel $c$ at layer $\ell$ in client $k$. For each source client $i \in \mathcal{N}_k$, we forward $x$ through client $i$ and obtain the feature map of its aligned channel $z_{i,c_i^{*}}^{(\ell)}(x)$, where $c_i^{*}$ is the channel from client $i$ aligned to target channel $c$ as determined in the alignment step. The aligned feature maps from source clients are aggregated as:
\begin{equation}
\label{equation_10}
\bar{z}_{k,c}^{(\ell)}(x)
=
\frac{1}{|\mathcal{N}_k|}
\sum_{i \in \mathcal{N}_k}
z_{i,c_i^{*}}^{(\ell)}(x),
\end{equation}
where aggregation involves no learnable parameters. The personalized feature map for client $k$ is then obtained via residual injection:
\begin{equation}
\hat{z}_{k,c}^{(\ell)}(x)=z_{k,c}^{(\ell)}(x)+\alpha \,\bar{z}_{k,c}^{(\ell)}(x),
\label{Eq: residual injection}
\end{equation}
where $\alpha$ controls the injection strength. Together, feature alignment and residual injection instantiate the representation-space operator $\mathcal{T}_k^{(\ell)}(\cdot)$ from Section~\ref{sec:problem}. This preserves the target model's original representation while incorporating aligned cross-domain knowledge. Importantly, the entire process operates in representation space without parameter modification or training.

\section{Theoretical Justification}
\label{sec:theory}
This section explains why representation-space alignment enables semantic-preserving and noise-reducing fusion in one-shot federated learning, by decomposing learned features into semantic components shared across domains and residual components induced by domain-specific and nuisance factors.
\subsection{Decomposition in Representation Space}
\label{sec:theory:view}

We assume that each data sample observed by client $k$ is generated as:
\begin{equation}
(Y,U)\sim P,\qquad
X_k = \gamma^\star(Y, D_k, U)
\end{equation}
where $Y$ denotes a semantic variable encoding task-relevant information, $D_k$ is a client-specific domain factor, and $U$ represents nuisance randomness.
The prediction target $T$ satisfies $T\perp D_k\mid Y$, implying that $Y$ is sufficient to determine the task semantics across domains.

Let $f(\cdot;\theta_k)$ denote the model trained on the local dataset of client $k$.
For a random input $X_k$, the feature extractor produces a random representation $\mathcal Z_k$.
We define the semantic component of the representation as the $L_2$-projection of $\mathcal Z_k$ onto the space of $\sigma(Y)$-measurable functions:
\begin{equation}
\label{ZKY}
\mathcal S_k = \mathbb E[\mathcal Z_k \mid Y]
\end{equation}

The remaining part of the representation is captured by the residual component:
\begin{equation}
\label{component}
\mathcal R_k = \mathcal Z_k - \mathcal S_k
\end{equation}
By the orthogonality property of conditional expectation, $\mathcal R_k$ satisfies:
\begin{equation}
\label{RY}
\mathbb E\!\left[\langle \mathcal R_k, \phi(Y)\rangle\right]=0 ,
\qquad \forall\, \phi(Y)\in L_2(\sigma(Y)),
\end{equation}
where $\phi(\cdot)$ denotes any arbitrary function of the semantic variable $Y$. Thus, $\mathcal R_k$ represents the part of the learned representation that cannot be explained solely by the task semantics encoded in $Y$. This component may arise from domain-specific factors associated with $D_k$, optimization noise, or other nuisance variations.

\subsection{From Semantic Alignment to Denoising Fusion}
\label{sec:theory:align_fuse}

We emphasize that client $k$ forwards its own local inputs through the feature extractors of other clients.
Specifically, for an input $X_k$, client $k$ leverages each source model $i \in \mathcal{N}_k$  to compute
\begin{equation}
\mathcal Z_i(X_k)=g(X_k;\theta_i), \qquad i\neq k
\end{equation}
without exposing raw data or labels and without updating any model parameters.

Since representations $\mathcal Z_i(X_k)$ are computed on the same target-domain input, their differences reflect both semantic discrepancies and domain-induced residual variations.
Using the Eq.~\eqref{component}, the representation distance naturally mixes semantic and residual components.
Therefore, representation alignment alone does not guarantee semantic consistency.

\textcolor{red}{Channels selected only from the high-alignment region, where$\qquad s\ge\tau$. We do not require the alignment score to serve as a globally exact semantic distance metric over all channel pairs. Instead, we assume local sufficient condition within the candidate region used for channel selection. Higher alignment scores are associated with smaller expected semantic discrepancy. Empirical results supporting this assumption are provided in Section~\ref{sec:assumption_validation}:
\begin{equation}
\label{score_17}
\mathbb E\!\left[
\|\mathcal S_i-\mathcal S_k\|_2^2
\mid
\mathrm{score}(\mathcal Z_i,\mathcal Z_k)=s
\right]
\le
b-a s,
\qquad s\ge\tau .
\end{equation}}
%%%%%%%%%%%%%%%%%%%%%%%%%%%%%%%%%%%%%%%
%%%%%%%%%%%%%%%%%%%%%%%%%%%%%%%%%%%%%%%
Our fusion process is implemented as follows,
\begin{equation}
\tilde{\mathcal Z}_k=(1-\alpha)\mathcal Z_k+\frac{\alpha}{m}\sum_{i\neq k}\,
\mathcal Z_i,
\end{equation}
where $m$ denotes the number of source representations selected according to the alignment criterion in Eq.~\eqref{CKAselect}. Using Eqs.~\eqref{ZKY}-\eqref{RY}, we decompose $\tilde{\mathcal Z}_k=\tilde{\mathcal S}_k+\tilde{\mathcal R}_k$. We have
\begin{equation}
\tilde{\mathcal S}_k-\mathcal S_k
=\alpha((\frac{1}{m}\sum_{i\neq k}\mathcal S_i)-\mathcal S_k\big),
\end{equation}
hence by Jensen's inequality,
\begin{equation}
\mathbb E\|\tilde{\mathcal S}_k-\mathcal S_k\|_2^2
\le \frac{\alpha^2}{m}\sum_{i\neq k}\,
\mathbb E\|\mathcal S_i-\mathcal S_k\|_2^2
\end{equation}
Under the sufficient condition in Eq.~\eqref{score_17}, higher alignment scores imply smaller expected semantic discrepancy, and thus the fused semantic component remains close to $\mathcal S_k$ in expectation.
%%%%%%%%%%%%%%%%%%%%%%%%%%%%
%%%%%%%%%%%%%%%%%%%%%%%%%%%%
We can similarly analyze the residual component ${\mathcal R}_k$, where
\begin{equation}
\begin{split}
\mathbb E[\|\tilde{\mathcal R}_k\|_2^2\mid Y]
&=(1-\alpha)^2\mathbb E[\|\mathcal R_k\|_2^2\mid Y] \\
&\quad+\frac{\alpha^2}{m^2}\,
\mathbb E\Big[\Big\|\sum_{i\neq k}\mathcal R_i\Big\|_2^2\mid Y\Big].
\end{split}
\end{equation}
where the cross term vanishes since $\mathbb E[\mathcal R_i\mid Y]=0$. If residuals across clients are not perfectly correlated given $Y$, then averaging strictly reduces the residual energy; in particular, $\tilde{\mathcal R}_k$ has smaller conditional energy than individual residuals, implying a denoising effect.

\subsection{Limitations of Parameter-Space Alignment}

The above analysis relies on the ability to establish semantic consistency across models through their responses to the same input.
In contrast, naive parameter-space similarity (e.g., Euclidean distance or direct averaging) is generally an unreliable proxy for semantic similarity due to permutation and scaling symmetries in neural network parametrizations. As a result, parameter-space aggregation lacks a well-defined mechanism to isolate semantic components from domain-specific or noisy variations, and therefore cannot replicate the denoising behavior enabled by representation-space alignment and fusion.

\begin{table*}[!ht]
\centering
\scriptsize
\caption{\textcolor{red}{Comparison of different methods under the few-shot setting on DomainNet, PACS, and Office-Home. 
Classification accuracy (\%) is reported as mean $\pm$ standard deviation over three independent runs where available, and the best results are highlighted in bold.}}
\label{as_exp}
\setlength{\tabcolsep}{2.6pt}
\newcommand{\accm}[2]{#1{\scriptsize$\pm$#2}}
\newcommand{\baccm}[2]{\textbf{#1{\scriptsize$\pm$#2}}}
\resizebox{\textwidth}{!}{
\begin{tabular}{c|c|ccc|cccccc}
\toprule
\multirow{2}{*}{Dataset} & \multirow{2}{*}{Domain}
& \multicolumn{3}{c|}{Non-OSFL Methods}
& \multicolumn{6}{c}{OSFL Methods} \\
\cmidrule(lr){3-5} \cmidrule(lr){6-11}
& 
& FedAvg & FDRL & VQCFL 
& DENSE & FOL & FedBiP & FedDEO & FedPFT & CRIP\\
\midrule

\multirow{7}{*}{DomainNet}
& C   
& \accm{73.12}{1.54} & \accm{81.49}{0.56} & \accm{79.11}{1.79}
& \accm{63.84}{2.51} & \accm{60.06}{25.01} & \accm{77.52}{0.67} & \accm{72.33}{1.26} & \accm{74.68}{0.56} & \baccm{86.90}{5.53}\\

& I   
& \accm{59.85}{1.51} & \accm{55.56}{0.88} & \accm{54.98}{1.85}
& \accm{52.87}{0.38} & \accm{52.30}{6.05} & \accm{60.94}{2.08} & \accm{57.39}{0.84} & \accm{58.62}{2.07} & \baccm{67.82}{2.07}\\

& P   
& \accm{63.77}{1.12} & \accm{74.98}{3.07} & \accm{73.90}{1.61}
& \accm{62.07}{0.97} & \accm{46.87}{9.45} & \accm{65.20}{0.78} & \accm{63.17}{1.05} & \accm{76.49}{0.07} & \baccm{84.61}{1.01}\\

& Q   
& \accm{16.26}{2.60} & \accm{72.62}{1.30} & \accm{26.53}{4.30}
& \accm{29.92}{1.62} & \accm{54.11}{17.26} & \accm{51.85}{3.24} & \accm{37.86}{2.47} & \accm{49.49}{1.32} & \baccm{92.00}{3.57}\\

& R   
& \accm{87.90}{0.09} & \baccm{96.99}{0.72} & \accm{94.90}{1.08}
& \accm{81.69}{1.14} & \accm{42.27}{13.20} & \accm{83.16}{0.60} & \accm{81.51}{1.03} & \accm{93.16}{0.66} & \accm{96.41}{2.08}\\

& S   
& \accm{68.07}{4.67} & \accm{76.35}{1.36} & \accm{77.55}{4.67}
& \accm{59.20}{2.12} & \accm{31.07}{3.81} & \accm{68.24}{0.78} & \accm{62.86}{1.61} & \accm{75.60}{1.43} & \baccm{78.99}{7.60}\\

\cmidrule(lr){2-11}
& Avg 
& \accm{61.49}{0.58} & \accm{76.33}{0.59} & \accm{67.83}{0.84}
& \accm{58.26}{1.33} & \accm{47.78}{4.93} & \accm{67.82}{0.56} & \accm{62.52}{1.56} & \accm{71.34}{0.38} & \baccm{84.46}{1.77}\\

\midrule
\midrule

\multirow{5}{*}{PACS}
& A   
& \accm{52.68}{3.22} & \accm{72.30}{1.32} & \accm{64.76}{1.99}
& \accm{44.64}{0.14} & \accm{80.83}{0.64} & \accm{53.26}{2.54} & \accm{49.89}{0.91} & \accm{66.47}{1.47} & \baccm{93.68}{0.92}\\

& C   
& \accm{68.27}{4.22} & \accm{73.61}{4.50} & \accm{69.51}{3.46}
& \accm{63.10}{1.47} & \accm{80.36}{0.67} & \accm{70.90}{2.97} & \accm{68.31}{1.41} & \accm{65.29}{1.72} & \baccm{92.88}{0.75}\\

& P   
& \accm{86.31}{1.03} & \baccm{95.03}{0.42} & \accm{87.01}{1.14}
& \accm{74.70}{0.81} & \accm{97.15}{0.88} & \accm{74.85}{1.36} & \accm{71.96}{0.56} & \accm{94.13}{0.63} & \accm{93.38}{0.43}\\

& S   
& \accm{31.25}{9.94} & \accm{63.06}{1.32} & \accm{42.95}{9.75}
& \accm{31.40}{2.06} & \accm{76.61}{2.64} & \accm{51.70}{1.69} & \accm{48.95}{1.34} & \accm{57.17}{0.42} & \baccm{92.01}{4.08}\\

\cmidrule(lr){2-11}
& Avg 
& \accm{59.63}{3.13} & \accm{76.00}{1.24} & \accm{66.06}{3.58}
& \accm{53.46}{1.62} & \accm{83.74}{0.63} & \accm{62.67}{0.45} & \accm{59.78}{1.07} & \accm{70.77}{0.49} & \baccm{94.24}{1.05}\\

\midrule
\midrule

\multirow{5}{*}{OfficeHome}
& A   
& \accm{54.48}{1.60} & \accm{62.41}{2.03} & \accm{58.57}{1.32}
& \accm{52.37}{0.96} & \baccm{65.29}{1.56} & \accm{55.41}{0.55} & \accm{49.37}{2.06} & \accm{61.18}{1.04} & \accm{62.55}{2.06}\\

& C   
& \accm{47.63}{1.08} & \accm{64.68}{0.83} & \accm{66.06}{1.43}
& \accm{46.24}{1.74} & \baccm{73.01}{1.26} & \accm{48.62}{0.42} & \accm{42.92}{0.81} & \accm{56.35}{0.26} & \accm{60.86}{1.74}\\

& P   
& \accm{73.94}{1.27} & \accm{84.01}{0.68} & \accm{81.98}{1.13}
& \accm{73.76}{2.07} & \baccm{85.66}{0.34} & \accm{76.63}{0.20} & \accm{73.81}{0.46} & \accm{77.33}{0.34} & \accm{81.01}{1.24}\\

& R   
& \accm{63.94}{0.56} & \accm{76.38}{0.61} & \accm{75.69}{0.46}
& \accm{61.86}{1.45} & \baccm{80.96}{1.83} & \accm{65.43}{0.96} & \accm{61.77}{0.51} & \accm{76.83}{1.00} & \accm{78.67}{1.19}\\

\cmidrule(lr){2-11}
& Avg 
& \accm{60.00}{0.88} & \accm{71.87}{0.50} & \accm{70.57}{0.32}
& \accm{58.55}{1.35} & \baccm{76.23}{0.51} & \accm{61.52}{0.39} & \accm{56.96}{1.71} & \accm{67.92}{0.43} & \accm{70.77}{1.14}\\

\bottomrule
\end{tabular}
}
\end{table*}

\section{Experiments and Analyses}
\begin{figure}[t]
    \centering
    \includegraphics[width=\columnwidth]{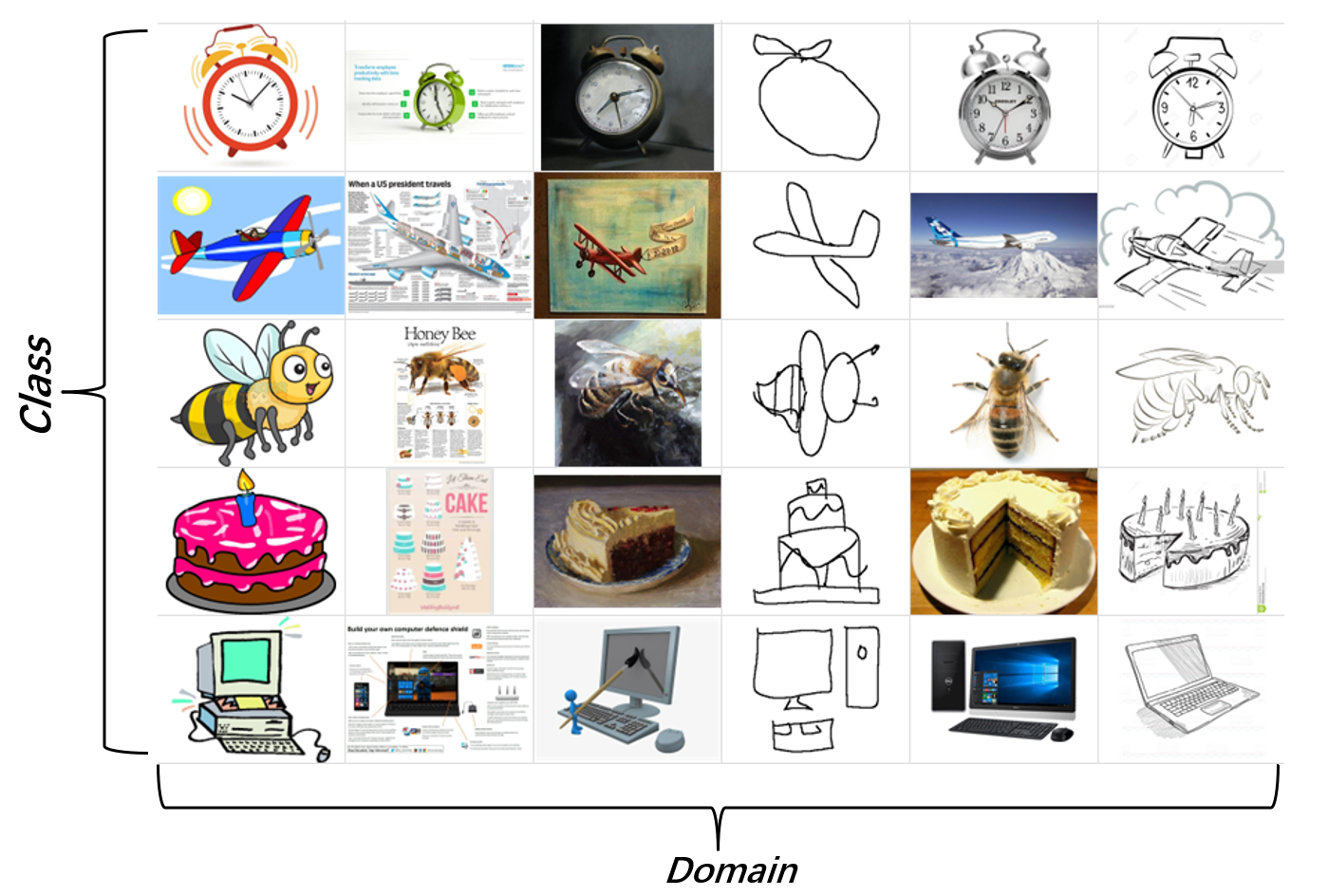}
    \caption{Illustration of domain heterogeneity in DomainNet. Each row corresponds to one semantic class, while each column represents a different domain. Although samples in the same row share the same class label, their visual appearances vary substantially across domains, including clipart, infograph, painting, quickdraw, real, and sketch. This highlights the severe cross-domain variation in style, texture, and visual abstraction, which makes parameter-space aggregation difficult and motivates representation-space personalization.}
    \label{fig:domainnet_heterogeneity}
\end{figure}

\subsection{Experimental Setup}
\paragraph{Domain-Heterogeneous Datasets}
We study the effectiveness of the proposed CRIP on three widely adopted domain-heterogeneous benchmarks, i.e., DomainNet~\cite{DomainNet}, PACS~\cite{PACS}, and Office-Home~\cite{office-home}. Example images from the domain-heterogeneous datasets are shown in Fig.~\ref{fig:domainnet_heterogeneity}. Specifically, DomainNet consists of six domains, Clipart (C), Infograph (I), Painting (P), Quickdraw (Q), Real (R), and Sketch (S), from which we select 10 categories. PACS includes four domains, namely Art (A), Cartoon (C), Photo (P), and Sketch (S), covering a total of 7 categories. Office-Home contains four domains, Art (A), Clipart (C), Product (P), and Real (R), with 65 categories selected for evaluation.
In all experiments involving our proposed method, each client corresponds to one domain, and the learning process is constrained to a single round of communication, which strictly follows the OSFL setting. We also simulate local data scarcity under heterogeneous data distributions. We follow FedBiP and adopt a few-shot setting, where each client is trained with 16 samples per class for DomainNet and PACS, and 8 samples per class for Office-Home.

\paragraph{Baselines}
\textcolor{red}{We compare CRIP with representative Muti-round federated learning methods and state-of-the-art OSFL approaches.}

\textcolor{red}{For Muti-round methods, we include the FedAvg baseline and two recent representation learning methods, FDRL\cite{FDRLTnnls} and VQCFL\cite{VQCFLTnnls}. Following a unified experimental protocol, all multi-round FL methods are trained for 100 communication rounds.}

For OSFL methods, we compare with DENSE, a data-free one-shot model distillation method; FOL, a one-shot personalized federated learning method; FedBiP and FedDEO, two diffusion-based OSFL methods; and FedPFT, a foundation-model-based one-shot federated learning method based on parametric feature transfer. These methods represent recent OSFL approaches based on model distillation, personalized learning, diffusion-based data generation, and foundation-model feature transfer.

Consistent with FedBiP, we adopt ResNet-18 pretrained on ImageNet as the backbone model. No public data or auxiliary labeled datasets are introduced for CRIP. All methods are evaluated independently on each target domain, and the average accuracy across domains is reported for each dataset. \textcolor{red}{Unless otherwise specified, all results are reported as the mean $\pm$ standard deviation over three independent runs with random seeds 42, 123, and 456.}

\subsection{Main Results}
Table~\ref{as_exp} summarizes the performance of different methods under the few-shot setting on DomainNet, PACS, and Office-Home. Overall, CRIP achieves the best average performance on DomainNet and PACS, reaching 84.46\% and 94.24\%, respectively, outperforming the strongest competing methods by 8.13 and 10.50 percentage points. On DomainNet, CRIP achieves the best performance on five out of six domains, with a particularly substantial gain on the highly shifted Quickdraw (Q) domain, where its accuracy reaches 92.00\%, compared with 72.62\% for the strongest baseline. Similarly, on PACS, CRIP obtains the best results on three out of four domains and demonstrates particularly strong performance on challenging domains such as Sketch (S). \textcolor{red}{Although FOL achieves the best average performance on Office-Home, its performance on DomainNet is considerably less stable across independent runs, exhibiting large standard deviations on several domains. In contrast, CRIP demonstrates a more favorable overall performance profile across the three heterogeneous benchmarks. These results indicate that directly operating in the representation space enables CRIP to effectively mitigate domain heterogeneity, particularly under substantial cross-domain distribution shifts.}

\begin{table*}[t]
\centering
\small
\caption{Accuracy improvement on PACS-Cartoon under different $\alpha$ values.}
\label{tab:cartoon_alpha}
\begin{tabular}{c|ccccccccc}
\toprule
$\alpha$ & 0.1 & 0.2 & 0.3 & 0.4 & 0.5 & 0.6 & 0.7 & 0.8 & 0.9 \\
\midrule
$\Delta$ Acc. & +0.0000 & -0.0026 & +0.0026 & +0.0004 & -0.0043 & -0.0175 & -0.0303 & -0.0516 & -0.0836 \\
\bottomrule
\end{tabular}
\end{table*}

We additionally evaluated the \textit{Cartoon} domain under different values of $\alpha$  in Eq. (\ref{Eq: residual injection}). As shown in Table.~\ref{tab:cartoon_alpha}, after adjusting the injection strength, the method does not suffer from severe degradation on this domain. In fact, the performance remains close to the baseline for smaller $\alpha$ values, and even shows slight improvement at $\alpha=\,$\textit{0.3} and $\alpha=\,$\textit{0.4}. These results suggest that the observed drop mainly comes from the mismatch between a globally fixed $\alpha$ and the domain-specific optimum for \textit{Cartoon}, rather than from an inherent failure of the proposed alignment-injection mechanism.

{\color{red}
\subsection{Validation of CKA-Based Alignment Assumption}
\label{sec:assumption_validation}

To validate the assumptions proposed in our theoretical analysis, we examine the relationship between CKA similarity and semantic correspondence in this section.

The channels exhibit different activations for data from different classes. 
Therefore, we use the class profile of a channel, defined as the distribution of its activation levels between classes, to characterize its class-level semantic preference.

Let $a^c(x)\in\mathbb{R}^{H\times W}$ denote the feature map of channel $c$ for input $x$.
We first obtain a scalar response $A^c$ by applying global average pooling to the feature map $a^c$:

\begin{equation}
A^c(x)
=
\frac{1}{HW}
\sum_{h=1}^{H}
\sum_{w=1}^{W}
a^c_{h,w}(x).
\end{equation}

For a dataset containing $Y$ classes, the class profile $D^c$ of channel $c$ is defined as
\begin{equation}
\mathbf D^c
=
\left[
\mu^c_1,
\mu^c_2,
\ldots,
\mu^c_Y
\right],
\end{equation}
where
\begin{equation}
\mu^c_j
=
\frac{1}{|\mathcal X_j|}
\sum_{x\in\mathcal X_j}
A^c(x)
\end{equation}
denotes the mean response of the channel $c$ to samples from class $j$.

Given a source model channel $c$ and a target model channel $c'$, we quantify their semantic correspondence by Pearson correlation between their $\mathbf D^c$ and $\mathbf D^{c'}$:

\begin{equation}
\label{eq:profile_similarity}
\operatorname{Sim}(c,c')
=
\frac{
\sum_{y=1}^{Y}
\left(\mu^c_y-\bar{\mu}^c\right)
\left(\mu^{c'}_y-\bar{\mu}^{c'}\right)
}{
\sqrt{
\sum_{y=1}^{Y}
\left(\mu^c_y-\bar{\mu}^c\right)^2
}
\sqrt{
\sum_{y=1}^{Y}
\left(\mu^{c'}_y-\bar{\mu}^{c'}\right)^2
}
},
\end{equation}
where
$\bar{\mu}^c=\frac{1}{Y}\sum_{y=1}^{Y}\mu^c_y$
and
$\bar{\mu}^{c'}=\frac{1}{Y}\sum_{y=1}^{Y}\mu^{c'}_y$
denote the mean class-wise responses of channels $c$ and $c'$, respectively.

Unlike $\operatorname{CKA}(c,c')$, which is computed only from feature representations, $\operatorname{Sim}(c,c')$ relies on the ground-truth labels of the samples, therefore, provides a proxy based on label for class level semantic correspondence.
A higher positive value indicates a stronger class-level semantic correspondence, whereas values near zero or negative values indicate weak or opposing correspondence, respectively.

If the CKA metric can capture the semantic correspondence between channels, then channel pairs with higher $\operatorname{CKA}(c,c')$ similarity should also exhibit higher $\operatorname{Sim}(c,c')$ values. 
To verify this assumption, we compute the complete $\operatorname{CKA}$ similarity matrix between the channels of the target model and those of the source model, and rank all channel pairs according to their $\operatorname{CKA}$ scores. 
For a given selection ratio $q$, let $\mathcal{T}_q$ and $\mathcal{B}_q$ denote the sets containing the top $q\%$ and bottom $q\%$ of all pair of channels ranked according to their scores $\operatorname{CKA}$, respectively. 

we compute the average $\operatorname{Sim}$ value over all included channel pairs. We also report Pearson $r$ and Spearman $\rho$ within each selected region. Pearson $r$ measures the linear association between CKA similarity and semantic correspondence, whereas Spearman $\rho$ measures their rank-order association. The reported mean and standard deviation of each metric are calculated across different experimental units. Finally, the $p$-value is obtained from a one-sided paired test over experimental units, testing whether the mean paired difference $M(\mathcal{T}_q) - M(\mathcal{B}_q)$ is significantly greater than zero.

The results in Table~\ref{tab:cka_region_top1} provide consistent empirical support for our theoretical assumption. 
Across all three datasets, the high-CKA region $\mathcal{T}_{1\%}$ consistently achieves larger profile similarity than the low-CKA region $\mathcal{B}_{1\%}$. 
This indicates that channel pairs with the highest CKA scores are much more likely to share similar semantic preference.

The same trend is also observed for both correlation metrics. 
Compared with the low-CKA region, the high-CKA region consistently shows larger Pearson $r$ and Spearman $\rho$, suggesting that within high-CKA regions, CKA similarity is more positively associated with semantic profile similarity. 

Importantly, this phenomenon is not limited to the most extreme Top $1\%$ channel pairs. 
As the selection ratio is progressively increased to $50\%$, the high-CKA regions still outperform the corresponding low-CKA regions across all datasets and metrics, with all paired tests yielding highly significant $p$-values.
{\color{green}We report the experimental results under additional selection ratios in Table I of the supplementary material.}
These results demonstrate that higher CKA scores are systematically associated with stronger semantic alignment between source and target channels. 
\begin{table*}[t]
\centering
\caption{
{\color{red}
Comparison between the Bottom $1\%$ and Top $1\%$ CKA regions.
For each experimental unit, channel pairs are ranked according to their
CKA scores. Results are reported as mean $\pm$ standard deviation across
experimental units. More detailed results under additional selection
ratios are provided in the supplementary material.
}
}
\label{tab:cka_region_top1}

\scriptsize
\setlength{\tabcolsep}{7pt}
\renewcommand{\arraystretch}{1.10}

\begin{tabular}{llcccc}
\toprule
Dataset
& Metric
& $\mathcal{B}_{1\%}$
& $\mathcal{T}_{1\%}$
& $\mathcal{T}_{1\%}-\mathcal{B}_{1\%}$
& $p$-value \\
\midrule

% ==================== DomainNet ====================

\multirow{3}{*}{DomainNet}
& Sim
& $-0.0035 \pm 0.0094$
& $0.1631 \pm 0.0990$
& $0.1666$
& $1.09 \times 10^{-27}$ \\

& $r$
& $-0.0017 \pm 0.0181$
& $0.1665 \pm 0.1094$
& $0.1682$
& $1.08 \times 10^{-25}$ \\

& $\rho$
& $-0.0018 \pm 0.0191$
& $0.1710 \pm 0.1259$
& $0.1728$
& $9.52 \times 10^{-23}$ \\

\midrule

% ==================== PACS ====================

\multirow{3}{*}{PACS}
& Sim
& $0.0045 \pm 0.0136$
& $0.1456 \pm 0.1085$
& $0.1411$
& $3.85 \times 10^{-9}$ \\

& $r$
& $0.0039 \pm 0.0217$
& $0.2036 \pm 0.1643$
& $0.1996$
& $6.37 \times 10^{-9}$ \\

& $\rho$
& $0.0043 \pm 0.0235$
& $0.1922 \pm 0.1612$
& $0.1879$
& $2.34 \times 10^{-8}$ \\

\midrule

% ==================== Office-Home ====================

\multirow{3}{*}{Office-Home}
& Sim
& $-0.0066 \pm 0.0033$
& $0.3954 \pm 0.0495$
& $0.4020$
& $6.16 \times 10^{-34}$ \\

& $r$
& $0.0056 \pm 0.0191$
& $0.4907 \pm 0.0197$
& $0.4852$
& $5.97 \times 10^{-45}$ \\

& $\rho$
& $0.0060 \pm 0.0206$
& $0.6466 \pm 0.0398$
& $0.6406$
& $4.98 \times 10^{-41}$ \\

\bottomrule
\end{tabular}
\end{table*}

We further visualize the relationship between CKA and class-profile similarity among channel pairs in the Top-$1\%$ CKA region. {\color{green}The detailed results are presented in Figs.1-3 of the supplementary material.}

The results show that the degree of semantic correspondence tends to increase as CKA similarity increases. This indicates that CKA similarity not only reflects similarity in the activation space, but is also associated with consistency in class-level semantic preferences.

Although the strength of this relationship varies across different domain pairs, the overall results support the theoretical assumption: CKA-based channel matching increases the likelihood of selecting semantically aligned channels.

}
\subsection{Ablation Study}
\begin{figure}[htbp]
  \vskip 0.0in
  \begin{center}
    \centerline{\includegraphics[width=\columnwidth]{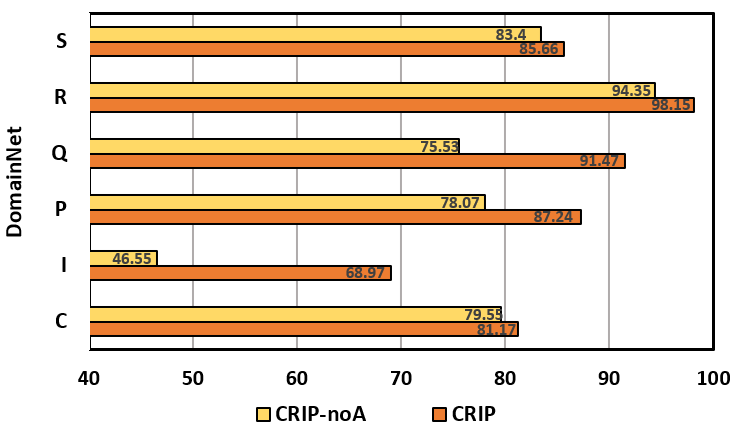}}
    \caption{Ablation study on DomainNet comparing CRIP with and without the feature alignment module. CRIP-noA denotes the variant without representation alignment.}
    \label{Ablation_study}
  \end{center}
  \vspace{-0.6cm}
\end{figure}
We compare \textbf{CRIP} with its variant \textbf{CRIP-noA}, which removes the alignment step, to evaluate the contribution of feature alignment. In CRIP-noA, residual injection directly averages representations from all clients without CKA-based channel selection. Fig.~\ref{Ablation_study} presents the classification accuracy per domain on DomainNet. The results show that removing the alignment module degrades performance in all domains.

These results are consistent with our theoretical analysis. In particular, statistical correlation alone does not necessarily imply semantic consistency. Our framework explicitly distinguishes the semantic component from the residual component. The latter captures domain-specific factors and nuisance variations. Therefore, the role of CKA in CRIP is not to guarantee perfect semantic equivalence. Instead, it serves as an alignment criterion for identifying channels that are more likely to be semantically transferable. As discussed in Section~\ref{sec:theory}, alignment by itself is insufficient to ensure semantic consistency; rather, under the sufficient condition in Eq.~\eqref{score_17}, higher alignment scores correspond to smaller expected semantic discrepancy.

From this perspective, the performance drop of CRIP-noA can be attributed to the lack of an alignment mechanism. This suggests that directly averaging unaligned representations is ineffective. The reason is that semantically mismatched channels and residual noise may be mixed across clients, thereby introducing representation interference. In contrast, after the alignment step selects more transferable channels, the averaging operation in Eq.~\eqref{equation_10} acts as a denoising operator: when residual components across clients are not perfectly correlated, averaging suppresses non-semantic statistical noise and reduces residual variance. Therefore, the proposed feature alignment is essential for enabling reliable residual injection and effective cross-client knowledge transfer under domain heterogeneity.

\subsection{Sensitivity Analysis}
We analyze two key factors that may affect the behavior of the proposed method. The first is the residual injection strength parameter $\alpha$, which controls the influence of cross-client aligned features in the personalized representation. The second is the choice of domain-specific samples $\mathcal{B}$ used during calibration, which determines the channel-wise CKA matching across clients.

\begin{figure}[tbp]
  \vskip 0.2in
  \begin{center}
    \centerline{\includegraphics[width=\columnwidth]{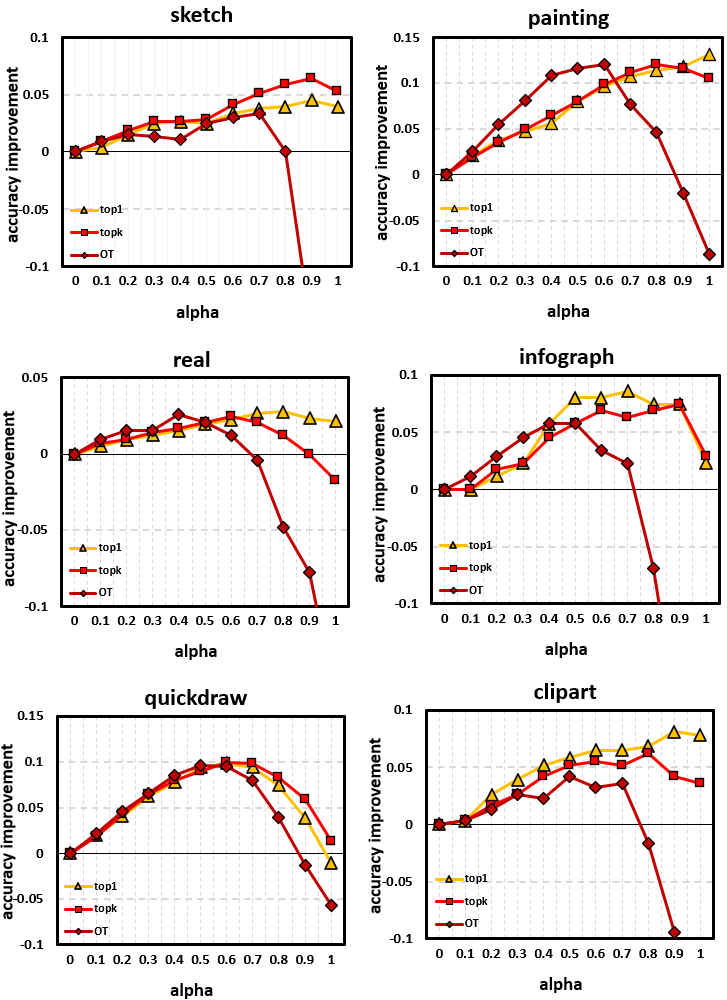}}
    \caption{Sensitivity analysis of the residual injection strength $\alpha$ on DomainNet. The figure shows classification accuracy per domain as $\alpha$ varies from $0$ to $1$.}
    \label{Sensitivity_Analysis}
  \end{center}
  \vspace{-0.6cm}
\end{figure}

Larger values of $\alpha$ increase influence from aligned features of other clients, and $\alpha=0$ corresponds to using only the target local model for prediction. Fig.~\ref{Sensitivity_Analysis} illustrates the changes in per-domain classification accuracy improvements on DomainNet as $\alpha$ increases from $0$ to $1$ under three different alignment strategies. 
Overall, moderate to large values of $\alpha$ consistently improve performance across most domains. These results validate the effectiveness of cross-client residual injection as a means of knowledge integration in representation space.

Although different domains exhibit varying sensitivity to $\alpha$, CRIP maintains stable performance over a wide range of $\alpha$ values without noticeable degradation. In contrast, the OT-based alignment strategy becomes less stable at larger $\alpha$ values. This suggests that effective cross-domain representation information is often concentrated in a subset of key channels. By comparison, overly balanced feature allocation may introduce noise. Taken together, these results indicate that the proposed feature alignment mechanism can effectively control the quality of injected features. By doing so, it enhances robustness in the presence of domain heterogeneity. Hence, in all experiments, we set $\alpha=0.8$ to provide a favorable trade-off between leveraging cross-client knowledge and preserving the discriminative power of local representations.

\begin{figure}[tbp]
  \vskip 0.2in
  \begin{center}
    \centerline{\includegraphics[width=\columnwidth]{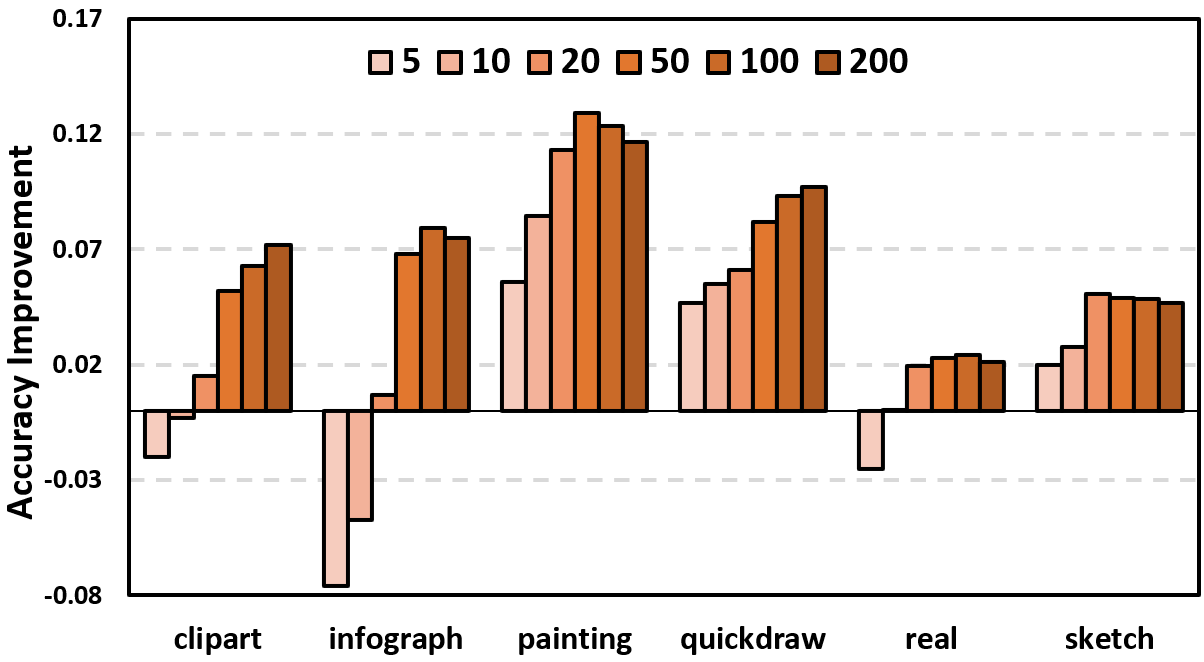}}
    \caption{Sensitivity analysis of domain-specific samples $\mathcal{B}$ on DomainNet. The figure shows accuracy improvement per domain as $\mathcal{B}$ varies from $5$ to $200$. Colors in the legend progress from light to dark, indicating increasing calibration sample size.}
    \label{Sensitivity_Analysis_beta}
  \end{center}
  \vspace{-0.6cm}
\end{figure}

We further analyze the effect of the calibration batch size used to construct the domain-specific calibration set $\mathcal{B}$. Fig.~\ref{Sensitivity_Analysis_beta} shows the per-domain accuracy improvement on DomainNet as the calibration batch size increases from 5 to 200. For each calibration batch size, we independently sample the calibration set five times and report the averaged results. Overall, a clear trend can be observed: increasing the calibration batch size generally leads to more reliable channel matching and larger performance gains across most domains.

The gains tend to saturate once the batch size becomes sufficiently large. For painting, the best performance is achieved around 50, followed by a slight decline at 100 and 200; similarly, real and sketch exhibit only marginal changes after 50. This indicates that once channel matching becomes sufficiently stable, further increasing the calibration batch size brings limited additional benefit.

Overall, these results show that CLIP is sensitive to the quality of calibration data, but remains robust once a moderate number of domain-specific samples is available.

To justify the use of \textbf{CRIP-K}, we analyzed the quantile statistics of the CKA similarity matrix. The distribution is highly skewed, where $q_p$ denotes the $p$-th quantile of the distribution of all pairwise CKA scores: $q_{0}=0.031653$, $q_{0.25}=0.235313$, $q_{0.5}=0.314525$, $q_{0.75}=0.409844$, $q_{0.9}=0.507017$, $q_{0.95}=0.568667$, $q_{0.99}=0.684875$, and the maximum value is $0.953321$. This indicates that most channel pairs fall into a moderate similarity range, while only a relatively small subset reaches clearly higher CKA scores.

Based on this observation, we introduced the \textbf{CRIP-K} variant to make use of these stronger statistical signals and to further examine the robustness of the method with respect to the number of selected channels. The sensitivity analysis shows that, after incorporating these high-ranking candidate channels, the model performance does not differ noticeably from using only the top-1 matched channel. This suggests that the method is reasonably robust to the channel-selection width within a certain range.

\subsection{Scalability Analysis}
\begin{figure}[htbp]
  \vskip 0.2in
  \begin{center}
    \centerline{\includegraphics[width=\columnwidth]{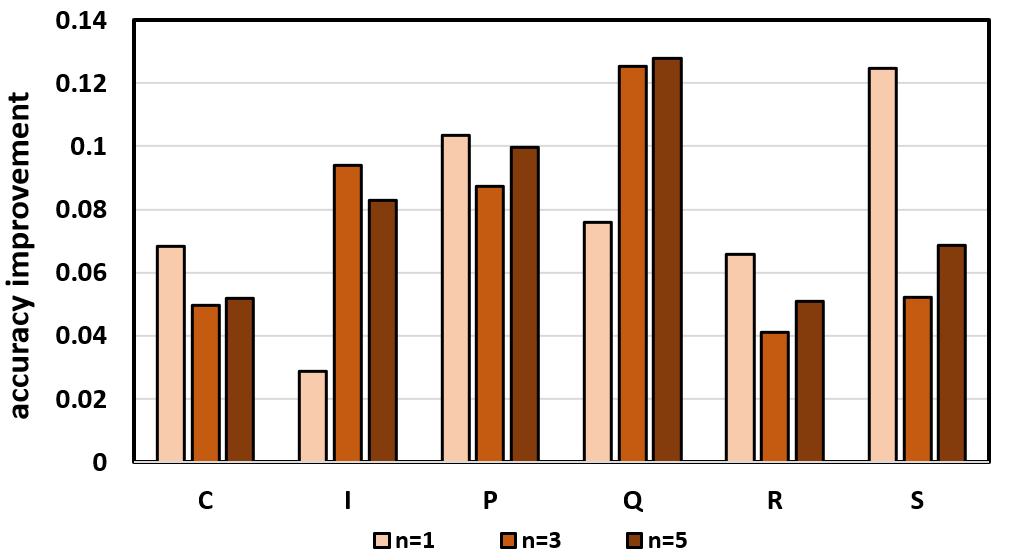}}
    \caption{Scalability analysis of the proposed method on DomainNet. The bar plots show the average accuracy improvement achieved by CRIP over local models for clients within each domain. Here, $n$ denotes the number of clients assigned to each domain.}
    \label{Scalability}
  \end{center}
  \vspace{-0.6cm}
\end{figure}
We conduct this scalability experiment to examine whether increasing the number of participating clients would introduce excessive feature mixing. Such mixing may potentially degrade performance in representation-space personalization. We vary the number of clients per domain n $\in$ {1, 3, 5} (i.e., up to 30 clients in total on DomainNet). As the number of clients per domain increases, a naive feature fusion strategy could suffer from noise accumulation and negative transfer.

However, the results in Fig.~\ref{Scalability} show that CRIP consistently yields positive accuracy gains across all domains even when more clients are involved. This indicates that CRIP is robust to client scale and does not suffer from performance degradation caused by mixing representations.

{\color{red}
\subsection{Privacy Risk Evaluation}
To assess the residual privacy risk associated with sharing feature extractors, we evaluate CRIP using shadow model membership inference attacks. The attacker uses auxiliary data drawn from the same domain distribution to train shadow models and an attack classifier based on intermediate representations.

\subsubsection{Shadow Model Training}
Both shadow model and client's model are initialized
from ImageNet pretrained ResNet-18 weights. The attacker replicates the client’s training pipeline using a disjoint subset from the target domain. The training configuration is kept identical to that of the target client: SGD is used with a learning rate of $0.001$, momentum of $0.9$, and weight decay of $5\times10^{-4}$; the loss function is cross-entropy loss. The shadow model is trained for a fixed number of $40$ epochs. For each domain in each dataset, we train a separate shadow model. All experiments are repeated using three random seeds.
\subsubsection{Representation Extraction}
The attacker extracts feature maps from \texttt{layer4.1.conv2} of both the shadow and target models using forward hooks, and applies global average pooling to obtain feature vectors. Features from the shadow model are used to train the attack classifier, whereas features from the target model are used for evaluation.

\subsubsection{Attack Classifier Training}
A binary classifier, denoted as \texttt{AttackMLP}, is trained using the feature vectors extracted from the shadow model. The input features are first standardized using z-score normalization, and the model is optimized using binary cross-entropy loss. The attack model relies only on intermediate representations.

\subsubsection{Membership Inference}
The attack model trained on the shadow model data is directly applied to the features extracted from the target model. It predicts a membership confidence score for each sample.

We report attack accuracy, AUC, TPR@1\%FPR, and attack advantage. Attack accuracy measures overall membership classification performance using a threshold of $0.5$, whereas AUC evaluates the ability to distinguish members from non-members across all thresholds. TPR@1\%FPR measures attack success under a strict low-false-positive setting. Attack advantage is defined as $\max_{\tau}[\mathrm{TPR}(\tau)-\mathrm{FPR}(\tau)]$, with higher values indicating greater privacy leakage.
\begin{table}[t]
\centering
\caption{
{\color{red}
Membership inference attack results. Results are reported as mean $\pm$ standard deviation.}
}
\label{tab:mia_results}

\scriptsize
\setlength{\tabcolsep}{2.8pt}
\renewcommand{\arraystretch}{1.08}

\begin{tabular}{@{}lcccc@{}}
\toprule
Dataset
& \makecell{Attack\\Acc.}
& AUC
& \makecell{TPR@1\%\\FPR}
& Adv. \\
\midrule

DomainNet
& $0.50 \pm 0.03$
& $0.50 \pm 0.03$
& $0.01 \pm 0.01$
& $0.06 \pm 0.03$ \\

Office-Home
& $0.58 \pm 0.04$
& $0.53 \pm 0.01$
& $0.02 \pm 0.01$
& $0.06 \pm 0.02$ \\

PACS
& $0.51 \pm 0.04$
& $0.51 \pm 0.05$
& $0.02 \pm 0.01$
& $0.09 \pm 0.05$ \\

\bottomrule
\end{tabular}
\end{table}

As shown in Table~\ref{tab:mia_results}, the attack accuracy and AUC remain close to the random guessing baseline overall. DomainNet and PACS produce values close to $0.5$. Although Office-Home exhibits a moderately higher attack accuracy, its AUC remains only $0.53$, indicating limited discriminative ability across decision thresholds. Moreover, TPR@1\%FPR remains low across all three datasets, ranging from $0.0108$ to $0.0231$, and the attack advantage remains below $0.10$ in all settings. {\color{green}Detailed results for each domain are reported in Tabel II of the supplementary material.} These results indicate that no substantial membership leakage is detected under the evaluated representation-level attack setting. However, they should not be interpreted as a guarantee of absolute privacy against all possible attacks.

}
\subsection{Denoising Effects}
\label{sec:exp:multi_domain}

\begin{table}[h]
\vspace{-0.2cm}
\centering
\caption{Average performance improvements (\%) under different domain combinations. For each target domain, the value is averaged over fusion results obtained by incorporating another single domain; \emph{All} denotes fusion with representations from all remaining domains.}
\label{tab:multi_domain_avg}
\resizebox{\columnwidth}{!}{%
\begin{tabular}{lccccccc}
\toprule
 & C & I & P & Q & R & S & All \\
\midrule
Avg & 7.1\% & 7.3\% & 7.5\% & 6.8\% & 7.8\% & 7.4\% & \textbf{8.0\%} \\
\bottomrule
\end{tabular}

}
\end{table}
To validate the theoretical justification in Section~\ref{sec:theory}, we compare fusion settings that incorporate either a single auxiliary domain or all remaining domains. The experimental results in Table~\ref{tab:multi_domain_avg} show that multi-domain fusion consistently achieves the largest average performance improvement across all settings. This finding is highly consistent with our theoretical analysis. Introducing a broader set of aligned representations in the representation space significantly enhances the denoising effect of the fusion process.
{\color{red}
\subsection{Inference Efficiency Analysis}
We evaluate the inference efficiency of CRIP on an NVIDIA GeForce RTX 4060 Laptop GPU with a batch size of $1$. {\color{green}The complete measurement protocol, component-wise runtime breakdown, and results for all values of $K$ are provided in Tables III and IV of the supplementary material.}

Under the current sequential implementation, both latency and FLOPs increase approximately linearly with the number of source models. Each additional source model introduces approximately $3.63$ GFLOPs and $2.7$--$3.5$ ms of latency. 

The component wise analysis further shows that most of the additional latency arises from feature extraction, whereas the target model forward pass with residual injection remains relatively stable. Because the source model forward passes are independent, their latency can potentially be reduced through parallel execution or source model selection. }
\section{Conclusion}
We propose CRIP, a feature-level framework for personalized OSFL. 
% It requires neither parameter aggregation nor additional training. Experiments demonstrate its effectiveness under severe domain shifts. 
In the future, CRIP can be extended to heterogeneous model architectures and multimodal settings.
% In this work, we propose CRIP, a personalized OSFL framework that operates entirely at the feature level. Using a small calibration set from the target domain, CRIP identifies channel-wise correspondences across client models via CKA similarity and integrates aligned features through residual connections, requiring no parameter aggregation or additional training.
% Extensive experiments on domain-heterogeneous benchmarks demonstrate that CRIP achieves state-of-the-art performance, with particularly significant gains under severe domain shifts.

% In future work, we plan to extend CRIP to more challenging settings, including heterogeneous model architectures and multimodal federated learning scenarios, where representation alignment across diverse feature spaces remains an open and important problem.

\bibliography{example_paper}
\bibliographystyle{IEEEtran}
% \begin{thebibliography}{1}
% \bibliographystyle{IEEEtran}
% \end{thebibliography}

\end{document}

\title{Supplementary Material}

\maketitle

% \section{Validation of CKA-Based Alignment Assumption}
\begin{table*}[t]
\centering
\caption{
Comparison between low-CKA and high-CKA regions under different selection ratios.
For each experimental unit and selection ratio $q$, the Bottom $q$ and Top $q$ channel pairs are selected according to their CKA scores.
Results are reported as mean $\pm$ standard deviation across experimental units.
}
\label{tab:cka_region_multi_ratio}

\scriptsize
\setlength{\tabcolsep}{5pt}
\renewcommand{\arraystretch}{1.08}

\begin{tabular}{lllcccc}
\toprule
Dataset
& Metric
& Ratio $q$
& $\mathcal{B}_q$
& $\mathcal{T}_q$
& $\mathcal{T}_q - \mathcal{B}_q$
& $p$-value \\
\midrule

% ==================== DomainNet ====================

\multirow{15}{*}{DomainNet}

& \multirow{5}{*}{Sim}
& 1\%
& $-0.0035 \pm 0.0094$
& $0.1631 \pm 0.0990$
& $0.1666$
& $1.09 \times 10^{-27}$ \\

&
& 10\%
& $-0.0021 \pm 0.0053$
& $0.0428 \pm 0.0233$
& $0.0450$
& $1.97 \times 10^{-29}$ \\

&
& 20\%
& $-0.0019 \pm 0.0046$
& $0.0255 \pm 0.0138$
& $0.0274$
& $3.13 \times 10^{-29}$ \\

&
& 30\%
& $-0.0017 \pm 0.0045$
& $0.0181 \pm 0.0098$
& $0.0198$
& $3.51 \times 10^{-29}$ \\

&
& 40\%
& $-0.0015 \pm 0.0044$
& $0.0140 \pm 0.0077$
& $0.0155$
& $8.46 \times 10^{-30}$ \\

&
& 50\%
& $-0.0013 \pm 0.0042$
& $0.0113 \pm 0.0064$
& $0.0125$
& $1.24 \times 10^{-29}$ \\

\cmidrule(lr){2-7}

&
\multirow{5}{*}{$r$}
& 1\%
& $-0.0017 \pm 0.0181$
& $0.1665 \pm 0.1094$
& $0.1682$
& $1.08 \times 10^{-25}$ \\

&
& 10\%
& $0.0019 \pm 0.0067$
& $0.0974 \pm 0.0610$
& $0.0956$
& $2.80 \times 10^{-26}$ \\

&
& 20\%
& $0.0013 \pm 0.0057$
& $0.0793 \pm 0.0485$
& $0.0780$
& $4.36 \times 10^{-27}$ \\

&
& 30\%
& $0.0014 \pm 0.0045$
& $0.0694 \pm 0.0426$
& $0.0679$
& $5.86 \times 10^{-27}$ \\

&
& 40\%
& $0.0015 \pm 0.0045$
& $0.0625 \pm 0.0387$
& $0.0610$
& $5.27 \times 10^{-27}$ \\

&
& 50\%
& $0.0020 \pm 0.0042$
& $0.0576 \pm 0.0357$
& $0.0556$
& $1.35 \times 10^{-26}$ \\

\cmidrule(lr){2-7}

&
\multirow{5}{*}{$\rho$}
& 1\%
& $-0.0018 \pm 0.0191$
& $0.1710 \pm 0.1259$
& $0.1728$
& $9.52 \times 10^{-23}$ \\

&
& 10\%
& $0.0018 \pm 0.0063$
& $0.0727 \pm 0.0445$
& $0.0710$
& $1.08 \times 10^{-26}$ \\

&
& 20\%
& $0.0011 \pm 0.0055$
& $0.0544 \pm 0.0316$
& $0.0533$
& $2.17 \times 10^{-28}$ \\

&
& 30\%
& $0.0013 \pm 0.0042$
& $0.0445 \pm 0.0257$
& $0.0432$
& $3.33 \times 10^{-28}$ \\

&
& 40\%
& $0.0012 \pm 0.0043$
& $0.0379 \pm 0.0221$
& $0.0367$
& $1.96 \times 10^{-28}$ \\

&
& 50\%
& $0.0017 \pm 0.0041$
& $0.0334 \pm 0.0193$
& $0.0317$
& $3.40 \times 10^{-28}$ \\

\midrule

% ==================== PACS ====================

\multirow{15}{*}{PACS}

& \multirow{5}{*}{Sim}
& 1\%
& $0.0045 \pm 0.0136$
& $0.1456 \pm 0.1085$
& $0.1411$
& $3.85 \times 10^{-9}$ \\

&
& 10\%
& $0.0023 \pm 0.0102$
& $0.0366 \pm 0.0280$
& $0.0344$
& $1.32 \times 10^{-8}$ \\

&
& 20\%
& $0.0022 \pm 0.0095$
& $0.0233 \pm 0.0186$
& $0.0211$
& $3.23 \times 10^{-8}$ \\

&
& 30\%
& $0.0020 \pm 0.0091$
& $0.0174 \pm 0.0147$
& $0.0154$
& $6.65 \times 10^{-8}$ \\

&
& 40\%
& $0.0020 \pm 0.0087$
& $0.0143 \pm 0.0128$
& $0.0123$
& $7.12 \times 10^{-8}$ \\

&
& 50\%
& $0.0021 \pm 0.0086$
& $0.0121 \pm 0.0115$
& $0.0100$
& $8.20 \times 10^{-8}$ \\

\cmidrule(lr){2-7}

&
\multirow{5}{*}{$r$}
& 1\%
& $0.0039 \pm 0.0217$
& $0.2036 \pm 0.1643$
& $0.1996$
& $6.37 \times 10^{-9}$ \\

&
& 10\%
& $0.0007 \pm 0.0066$
& $0.0923 \pm 0.0745$
& $0.0916$
& $3.91 \times 10^{-9}$ \\

&
& 20\%
& $0.0002 \pm 0.0053$
& $0.0693 \pm 0.0557$
& $0.0692$
& $2.80 \times 10^{-9}$ \\

&
& 30\%
& $-0.0002 \pm 0.0041$
& $0.0584 \pm 0.0465$
& $0.0586$
& $1.85 \times 10^{-9}$ \\

&
& 40\%
& $-0.0002 \pm 0.0043$
& $0.0508 \pm 0.0406$
& $0.0510$
& $1.49 \times 10^{-9}$ \\

&
& 50\%
& $0.0001 \pm 0.0041$
& $0.0458 \pm 0.0366$
& $0.0457$
& $1.21 \times 10^{-9}$ \\

\cmidrule(lr){2-7}

&
\multirow{5}{*}{$\rho$}
& 1\%
& $0.0043 \pm 0.0235$
& $0.1922 \pm 0.1612$
& $0.1879$
& $2.34 \times 10^{-8}$ \\

&
& 10\%
& $0.0013 \pm 0.0066$
& $0.0566 \pm 0.0436$
& $0.0553$
& $1.31 \times 10^{-9}$ \\

&
& 20\%
& $0.0004 \pm 0.0049$
& $0.0401 \pm 0.0304$
& $0.0397$
& $7.39 \times 10^{-10}$ \\

&
& 30\%
& $-0.0002 \pm 0.0039$
& $0.0329 \pm 0.0247$
& $0.0331$
& $2.64 \times 10^{-10}$ \\

&
& 40\%
& $-0.0001 \pm 0.0042$
& $0.0276 \pm 0.0210$
& $0.0277$
& $1.75 \times 10^{-10}$ \\

&
& 50\%
& $0.0003 \pm 0.0041$
& $0.0246 \pm 0.0189$
& $0.0243$
& $1.49 \times 10^{-10}$ \\

\midrule

% ==================== OfficeHome ====================

\multirow{15}{*}{OfficeHome}

& \multirow{5}{*}{Sim}
& 1\%
& $-0.0066 \pm 0.0033$
& $0.3954 \pm 0.0495$
& $0.4020$
& $6.16 \times 10^{-34}$ \\

&
& 10\%
& $-0.0062 \pm 0.0023$
& $0.0970 \pm 0.0149$
& $0.1032$
& $6.60 \times 10^{-31}$ \\

&
& 20\%
& $-0.0058 \pm 0.0021$
& $0.0544 \pm 0.0088$
& $0.0602$
& $2.47 \times 10^{-30}$ \\

&
& 30\%
& $-0.0056 \pm 0.0020$
& $0.0367 \pm 0.0063$
& $0.0423$
& $4.46 \times 10^{-30}$ \\

&
& 40\%
& $-0.0055 \pm 0.0019$
& $0.0269 \pm 0.0048$
& $0.0324$
& $3.53 \times 10^{-30}$ \\

&
& 50\%
& $-0.0053 \pm 0.0020$
& $0.0207 \pm 0.0040$
& $0.0260$
& $3.78 \times 10^{-30}$ \\

\cmidrule(lr){2-7}

&
\multirow{5}{*}{$r$}
& 1\%
& $0.0056 \pm 0.0191$
& $0.4907 \pm 0.0197$
& $0.4852$
& $5.97 \times 10^{-45}$ \\

&
& 10\%
& $0.0032 \pm 0.0053$
& $0.3198 \pm 0.0196$
& $0.3166$
& $4.44 \times 10^{-43}$ \\

&
& 20\%
& $0.0034 \pm 0.0038$
& $0.2837 \pm 0.0209$
& $0.2803$
& $7.11 \times 10^{-41}$ \\

&
& 30\%
& $0.0037 \pm 0.0041$
& $0.2647 \pm 0.0220$
& $0.2610$
& $2.08 \times 10^{-39}$ \\

&
& 40\%
& $0.0027 \pm 0.0043$
& $0.2520 \pm 0.0230$
& $0.2493$
& $4.51 \times 10^{-38}$ \\

&
& 50\%
& $0.0027 \pm 0.0044$
& $0.2430 \pm 0.0237$
& $0.2402$
& $1.06 \times 10^{-36}$ \\

\cmidrule(lr){2-7}

&
\multirow{5}{*}{$\rho$}
& 1\%
& $0.0060 \pm 0.0206$
& $0.6466 \pm 0.0398$
& $0.6406$
& $4.98 \times 10^{-41}$ \\

&
& 10\%
& $0.0029 \pm 0.0056$
& $0.2532 \pm 0.0415$
& $0.2503$
& $6.35 \times 10^{-29}$ \\

&
& 20\%
& $0.0024 \pm 0.0040$
& $0.1858 \pm 0.0296$
& $0.1833$
& $2.03 \times 10^{-29}$ \\

&
& 30\%
& $0.0025 \pm 0.0043$
& $0.1514 \pm 0.0236$
& $0.1488$
& $7.86 \times 10^{-30}$ \\

&
& 40\%
& $0.0014 \pm 0.0046$
& $0.1287 \pm 0.0200$
& $0.1273$
& $8.76 \times 10^{-30}$ \\

&
& 50\%
& $0.0009 \pm 0.0049$
& $0.1126 \pm 0.0171$
& $0.1117$
& $1.12 \times 10^{-29}$ \\

\bottomrule

\end{tabular}
\end{table*}

\begin{figure*}[htbp]
  \vskip 0.0in
  \begin{center}
    \centerline{\includegraphics[width=\textwidth]{domainnet_top1.pdf}}
    \caption{
    {\color{red}
    High-CKA channel-pair visualization on DomainNet. 
    Rows correspond to target domains, and columns correspond to source domains.}
    }
    \label{assumption_validation_domainnet}
  \end{center}
  \vspace{-0.6cm}
\end{figure*}
\begin{figure*}[htbp]
  \vskip 0.0in
  \begin{center}
    \centerline{\includegraphics[width=\textwidth]{officehome_top1.pdf}}
    \caption{
    {\color{red}
    High-CKA channel-pair visualization on OfficeHome. 
    Rows correspond to target domains, and columns correspond to source domains.}
    }
    \label{assumption_validation_officehome}
  \end{center}
  \vspace{-0.6cm}
\end{figure*}

\begin{figure*}[htbp]
  \vskip 0.0in
  \begin{center}
    \centerline{\includegraphics[width=\textwidth]{pacs_top1.pdf}}
    \caption{
    {\color{red}
    High-CKA channel-pair visualization on PACS. 
    Rows correspond to target domains, and columns correspond to source domains.}
    }
    \label{assumption_validation_pacs}
  \end{center}
  \vspace{-0.6cm}
\end{figure*}
% \section{Statistical Significance Tests}
% \section{Additional CKA Validation Results}
% \section{Privacy Attack Details}
% \begin{table*}[t]
% \centering
% \caption{
% {\color{red}
% Membership inference attack results. Lower values indicate lower privacy leakage.}}
% \label{tab:mia_results}
% \begin{tabular}{llcccc}
% \toprule
% Dataset  & Attack Acc. $\downarrow$ & AUC $\downarrow$ & TPR@1\%FPR $\downarrow$ & Adv. $\downarrow$ \\
% \midrule
% domainnet  & $0.4983 \pm 0.0346$ & $0.5036 \pm 0.0296$ & $0.0108 \pm 0.0121$ & $0.0620 \pm 0.0337$ \\
% officehome  & $0.5765 \pm 0.0432$ & $0.5250 \pm 0.0139$ & $0.0168 \pm 0.0081$ & $0.0618 \pm 0.0179$ \\
% pacs  & $0.5097 \pm 0.0425$ & $0.5104 \pm 0.0465$ & $0.0231 \pm 0.0140$ & $0.0923 \pm 0.0490$ \\
% \bottomrule
% \end{tabular}
% \end{table*}

\begin{table*}[t]
\centering
\caption{
Membership inference attack results per target domain (mean $\pm$ std across 3 seeds).
Lower values indicate lower privacy leakage.}
\label{tab:mia_results}
\begin{tabular}{llcccc}
\toprule
Dataset & Domain & Attack Acc. $\downarrow$ & AUC $\downarrow$ & TPR@1\%FPR $\downarrow$ & Memb. Adv. $\downarrow$ \\
\midrule

\multirow{7}{*}{DomainNet}
& clipart    & $0.4925 \pm 0.0152$ & $0.5033 \pm 0.0164$ & $0.0063 \pm 0.0063$ & $0.0623 \pm 0.0150$ \\
& infograph  & $0.4752 \pm 0.0707$ & $0.4834 \pm 0.0172$ & $0.0087 \pm 0.0100$ & $0.0420 \pm 0.0259$ \\
& painting   & $0.5083 \pm 0.0153$ & $0.5048 \pm 0.0412$ & $0.0125 \pm 0.0125$ & $0.0708 \pm 0.0382$ \\
& quickdraw  & $0.5000 \pm 0.0000$ & $0.5071 \pm 0.0120$ & $0.0000 \pm 0.0000$ & $0.0438 \pm 0.0272$ \\
& real       & $0.5106 \pm 0.0518$ & $0.5061 \pm 0.0380$ & $0.0271 \pm 0.0157$ & $0.0739 \pm 0.0409$ \\
& sketch     & $0.5031 \pm 0.0289$ & $0.5169 \pm 0.0518$ & $0.0104 \pm 0.0096$ & $0.0792 \pm 0.0546$ \\
\cmidrule(lr){2-6}
& \textbf{Avg.} & $0.4983 \pm 0.0346$ & $0.5036 \pm 0.0296$ & $0.0108 \pm 0.0121$ & $0.0620 \pm 0.0337$ \\

\midrule

\multirow{5}{*}{PACS}
& art\_painting & $0.5030 \pm 0.0418$ & $0.5041 \pm 0.0789$ & $0.0238 \pm 0.0137$ & $0.1042 \pm 0.0670$ \\
& cartoon      & $0.5119 \pm 0.0526$ & $0.5231 \pm 0.0503$ & $0.0179 \pm 0.0089$ & $0.1071 \pm 0.0540$ \\
& photo        & $0.4940 \pm 0.0624$ & $0.4885 \pm 0.0400$ & $0.0179 \pm 0.0155$ & $0.0595 \pm 0.0573$ \\
& sketch       & $0.5298 \pm 0.0246$ & $0.5259 \pm 0.0109$ & $0.0327 \pm 0.0185$ & $0.0982 \pm 0.0155$ \\
\cmidrule(lr){2-6}
& \textbf{Avg.} & $0.5097 \pm 0.0425$ & $0.5104 \pm 0.0465$ & $0.0231 \pm 0.0140$ & $0.0923 \pm 0.0490$ \\

\midrule

\multirow{5}{*}{OfficeHome}
& Art         & $0.6357 \pm 0.0201$ & $0.5195 \pm 0.0202$ & $0.0233 \pm 0.0069$ & $0.0638 \pm 0.0221$ \\
& Clipart     & $0.5759 \pm 0.0244$ & $0.5365 \pm 0.0048$ & $0.0119 \pm 0.0036$ & $0.0726 \pm 0.0142$ \\
& Product     & $0.5606 \pm 0.0295$ & $0.5280 \pm 0.0130$ & $0.0227 \pm 0.0063$ & $0.0677 \pm 0.0144$ \\
& Real World  & $0.5336 \pm 0.0023$ & $0.5160 \pm 0.0096$ & $0.0093 \pm 0.0047$ & $0.0432 \pm 0.0100$ \\
\cmidrule(lr){2-6}
& \textbf{Avg.} & $0.5765 \pm 0.0432$ & $0.5250 \pm 0.0139$ & $0.0168 \pm 0.0081$ & $0.0618 \pm 0.0179$ \\

\bottomrule
\end{tabular}
\end{table*}

\begin{table*}[t]
\centering
\caption{
{\color{red}
Inference overhead of CRIP compared with the single-model baseline.
Latency is reported as mean $\pm$ standard deviation over target domains
and random seeds. It is measured with a batch size of $1$ using GPU event
timing. FLOPs are counted over the complete forward pass, including both
the source and target models.}
}
\label{tab:inference_overhead}

\begin{tabular}{lcrrcc}
\toprule
Dataset
& $K$
& Latency (ms)
& FLOPs (G)
& Peak Memory (MB)
& Overhead \\
\midrule

\multirow{6}{*}{DomainNet}
& 0 & $3.1 \pm 1.1$  & $3.63$  & 1165 & $1.0\times$ \\
& 1 & $7.1 \pm 2.5$  & $7.25$  & 1178 & $2.3\times$ \\
& 2 & $11.3 \pm 3.8$ & $10.88$ & 1190 & $3.6\times$ \\
& 3 & $14.1 \pm 5.2$ & $14.51$ & 1202 & $4.5\times$ \\
& 4 & $17.4 \pm 6.6$ & $18.14$ & 1214 & $5.5\times$ \\
& 5 & $20.8 \pm 7.7$ & $21.76$ & 1227 & $6.7\times$ \\

\midrule

\multirow{4}{*}{Office-Home}
& 0 & $2.6 \pm 0.6$  & $3.63$  & 1295 & $1.0\times$ \\
& 1 & $5.6 \pm 1.4$  & $7.25$  & 1307 & $2.2\times$ \\
& 2 & $8.9 \pm 2.3$  & $10.88$ & 1319 & $3.4\times$ \\
& 3 & $10.9 \pm 2.5$ & $14.51$ & 1332 & $4.2\times$ \\

\midrule

\multirow{4}{*}{PACS}
& 0 & $2.4 \pm 0.4$  & $3.63$  & 1186 & $1.0\times$ \\
& 1 & $5.0 \pm 0.3$  & $7.25$  & 1199 & $2.1\times$ \\
& 2 & $8.1 \pm 1.3$  & $10.88$ & 1211 & $3.3\times$ \\
& 3 & $10.5 \pm 1.7$ & $14.51$ & 1223 & $4.3\times$ \\

\bottomrule
\end{tabular}
\end{table*}

% Table~\ref{tab:inference_overhead} summarizes the end-to-end inference cost of CRIP. Both latency and FLOPs increase approximately linearly with the number of source models $K$. Each additional source model requires one additional feature-extractor forward pass and increases the total computation by approximately $3.63$~GFLOPs. Under the current sequential implementation, the average latency increase per additional source model is approximately $2.7$--$3.5$~ms across the
% three datasets. Nevertheless, the absolute latency remains between $10.5$ and $14.1$~ms per image under the evaluated hardware configuration.

% The measured GPU memory growth remains modest. On DomainNet, peak memory increases from $1165$~MB at $K{=}0$ to $1227$~MB at $K{=}5$, corresponding to an increase of $62$~MB, or approximately $5.3\%$ of the baseline memory footprint. Similar trends are observed on Office-Home and PACS, where each additional source model increases the measured peak memory by approximately $12$--$13$~MB. These results indicate that the memory overhead grows slow.

% The source model forward passes are mutually independent. Therefore, with sufficient parallel hardware, they can be executed concurrently to reduce feature extraction latency. Such parallel execution does not reduce the total FLOPs and may increase peak memory usage, resulting in a trade-off between inference latency and computational resources.

\begin{table}[t]
\centering
\caption{
{\color{red}
Component-wise latency breakdown of CRIP inference with a batch size of
$1$. ``Extract'' denotes the cumulative latency of the $K$ source-model
forward passes, while ``Inject'' denotes the target-model forward pass
with residual injection. FLOPs are measured over the complete CRIP
forward pass, including both the source and target models.}
}
\label{tab:inference_breakdown}

\begin{tabular}{lcccc}
\toprule
Dataset
& $K$
& Extract (ms)
& Inject (ms)
& FLOPs (G) \\
\midrule

\multirow{5}{*}{DomainNet}
& 1 & $3.4 \pm 1.2$  & $3.8 \pm 1.3$ & $7.25$  \\
& 2 & $6.1 \pm 2.4$  & $4.1 \pm 1.5$ & $10.88$ \\
& 3 & $9.8 \pm 3.6$  & $4.3 \pm 1.5$ & $14.51$ \\
& 4 & $12.3 \pm 5.2$ & $4.3 \pm 1.6$ & $18.14$ \\
& 5 & $15.2 \pm 5.7$ & $4.6 \pm 1.5$ & $21.76$ \\

\midrule

\multirow{3}{*}{Office-Home}
& 1 & $2.5 \pm 0.6$ & $3.1 \pm 0.9$ & $7.25$  \\
& 2 & $5.2 \pm 1.6$ & $3.1 \pm 0.6$ & $10.88$ \\
& 3 & $7.1 \pm 1.4$ & $3.3 \pm 0.9$ & $14.51$ \\

\midrule

\multirow{3}{*}{PACS}
& 1 & $2.1 \pm 0.2$ & $2.7 \pm 0.3$ & $7.25$  \\
& 2 & $4.3 \pm 0.3$ & $2.9 \pm 0.3$ & $10.88$ \\
& 3 & $6.7 \pm 1.3$ & $3.1 \pm 0.5$ & $14.51$ \\

\bottomrule
\end{tabular}
\end{table}

% Table~\ref{tab:inference_breakdown} decomposes CRIP's inference latency into feature extraction and target injection. Because the two components are timed separately, their sum may differ slightly from the end-to-end latency.

% Under the current sequential implementation, feature extraction accounts for most of the additional computational cost. In contrast, the injection remains stable. On DomainNet, it increases from $3.8$~ms at $K{=}1$ to $4.6$~ms at $K{=}5$, and similar trends are observed on Office-Home and PACS. This confirms that the dominant inference overhead arises from feature extraction rather than from residual injection itself.

% Overall, CRIP's inference cost is predictable and can be decomposed into distinct computational stages. Its cost grows approximately linearly with the number of source models. There are two potential deployment strategies. First, independent source model forward passes can be parallelized across multiple GPUs. With sufficient parallel resources, this can reduce latency from linear growth with $K$ to nearly constant. Second, inference can be restricted to a subset of the most relevant source models identified during calibration. This could reduce FLOPs and latency, potentially at the cost of some accuracy. Larger scale deployments with more source models may therefore benefit from parallel execution, source model selection, or a combination of both.

\bibliographystyle{IEEEtran}
% \bibliography{references}